\documentclass[12pt]{article}

\usepackage{amsfonts}
\usepackage{color}
\usepackage{rawfonts}
\usepackage{amsfonts}

\input{epsf} 

\newtheorem{assumption}{Assumption}
\newtheorem{definition}{Definition}[section]
\newtheorem{example}{Example}[section]
\newtheorem{remark}{Remark}[section]
\newtheorem{lemma}{Lemma}[section]
\newtheorem{theorem}[lemma]{Theorem}
\newtheorem{corollary}[lemma]{Corollary}

\begin{document}

\title{\Large \bf Deriving a Stationary Dynamic Bayesian Network from a Logic Program
with Recursive Loops\thanks{A preliminary version appears in
the 15th International Conference on Inductive Logic Programming.}}

\author{Yi-Dong Shen\\
{\small  Laboratory of Computer Science, Institute of Software}\\
{\small Chinese Academy of Sciences, Beijing 100080, China}\\
{\small Email: ydshen@ios.ac.cn}\\[.1in]
Qiang Yang\\
{\small Department of Computing Science, Hong Kong University of Science and Technology}\\
{\small Hong Kong, China}\\ 
{\small Email: qyang@cs.ust.hk}\\[.1in]
Jia-Huai You and Li-Yan Yuan\\
{\small  Department of Computing Science, University of Alberta}\\
{\small  Edmonton, Alberta, Canada T6G 2H1}\\
{\small  Email: \{you, yuan\}@cs.ualberta.ca}
}

\date{}

\maketitle             

\begin{abstract} 
Recursive loops in a logic program present a challenging problem to the PLP framework. 
On the one hand, they loop forever so that the PLP backward-chaining inferences would never
stop. On the other hand, they generate cyclic influences, which are disallowed in
Bayesian networks. Therefore, in existing PLP approaches logic programs with recursive loops 
are considered to be problematic and thus are
excluded. In this paper, we propose an approach that 
makes use of recursive loops to build a stationary dynamic Bayesian network.
Our work stems from an observation that recursive 
loops in a logic program imply a time sequence and thus can be used to
model a stationary dynamic Bayesian network without using explicit time parameters. 
We introduce a Bayesian knowledge base with logic clauses
of the form $A \leftarrow A_1,...,A_l, true, Context, Types$,
which naturally represents
the knowledge that the $A_i$s have direct influences on $A$ in 
the context $Context$ under 
the type constraints $Types$. 
We then use the well-founded model of a logic program
to define the direct influence relation
and apply SLG-resolution to compute the space of random variables
together with their parental connections. 
We introduce a novel notion of influence clauses, 
based on which a declarative semantics for a Bayesian knowledge base
is established and algorithms for building
a two-slice dynamic Bayesian network from a logic program are developed. \\[.1in]
\end{abstract}

\section{Introduction}
\label{int}
Probabilistic logic programming (PLP) 
is a framework that extends the expressive power of
Bayesian networks with first-order logic \cite{ngo-had97,poole93}.
The core of the PLP framework is a backward-chaining procedure, which
generates a Bayesian network graphic structure from a logic program in a way quite like 
query evaluation in logic programming.
Therefore, existing PLP methods use a slightly 
adapted {\em SLD-} or {\em SLDNF-resolution} \cite{Ld87}
as the backward-chaining procedure. 

Recursive loops in a logic program are SLD-derivations of the form
\begin{equation}
\label{loop1} 
A_1\leftarrow ... \leftarrow A_2 \ ...\leftarrow A_3 \ ... 
\end{equation}
where for any $i\geq 1$, $A_i$ is the same as $A_{i+1}$ up to 
variable renaming.\footnote{The {\em left-most} computation rule \cite{Ld87}
is assumed in this paper.} Such loops present a challenging problem to the PLP framework. 
On the one hand, they loop forever so that the PLP backward-chaining inferences would never
stop. On the other hand, they may generate cyclic influences, which are disallowed in
Bayesian networks. 

Two representative
approaches have been proposed to avoid recursive loops. The first one is by
Ngo and Haddawy \cite{ngo-had97} and 
Kersting and De Raedt \cite{Kersting2000}, who restrict
to considering only acyclic logic programs \cite{apt91}.
The second approach, 
proposed by Glesner and Koller \cite{glesner95}, 
uses explicit time parameters to avoid occurrence of recursive loops. 
It enforces acyclicity using time 
parameters in the way that every predicate has a time argument such that the time argument
in the clause head is at least one time step later than the time arguments
of the predicates in the clause body. 
In this way, each predicate $p(X)$ is changed to $p(X, T)$ and each clause 
$p(X)\leftarrow q(X)$
is rewritten into 
$p(X,T1)\leftarrow T2=T1-1,q(X,T2)$,
where $T$, $T1$ and $T2$ are time parameters. 

In this paper, we propose a solution to the problem of 
recursive loops under the PLP framework.
Our method is not restricted to acyclic logic programs, nor does it rely on explicit
time parameters. Instead, it makes use of recursive
loops to derive a stationary dynamic Bayesian network. 
We will make two novel contributions. First, 
we introduce the {\em well-founded} semantics \cite{VRS91} of logic programs to 
the PLP framework; in particular, we use the well-founded model of a logic program
to define the direct influence relation
and apply {\em SLG-resolution} \cite{chen96} 
(or {\em SLTNF-resolution} \cite{shen004})
to make the backward-chaining inferences. As a result, termination
of the PLP backward-chaining process is guaranteed. 
Second, we observe that under the PLP framework recursive loops (cyclic influences) 
define feedbacks, thus implying a time sequence. 
For instance, the clause
$aids(X)\leftarrow aids(Y), contact(X,Y)$
introduces recursive loops
\[aids(X) \leftarrow aids(Y) \ ... \leftarrow aids(Y1) \ ...\] 
Together with some other clauses in a logic program, 
these recursive loops may generate
cyclic influences of the form
\[aids(p1)\leftarrow ... \leftarrow aids(p1) \ ... \leftarrow aids(p1) \ ...\]
Such cyclic influences represent feedback connections, i.e.,  
that $p1$ {\em is} infected with aids (in the current time slice $t$) depends on 
whether $p1$ {\em was} infected with aids earlier (in the last time slice $t-1$).
Therefore, recursive loops of form (\ref{loop1}) imply a time sequence of the form
\begin{equation}
\label{time-sequence} 
A\underbrace{\leftarrow ... \leftarrow}_t A\underbrace{\ ... \leftarrow}_{t-1} A\underbrace{\ ... \leftarrow}_{t-2} A\ ...
\end{equation}
where $A$ is a ground instance of $A_1$.
It is this observation that leads us to viewing a logic program 
with recursive loops as a special temporal model. 
Such a temporal model corresponds to a stationary dynamic Bayesian network and 
thus can be compactly represented as a two-slice dynamic Bayesian network.

The paper is structured as follows. In Section 2, we review some 
concepts concerning Bayesian networks and logic programs. In Section 3, we 
introduce a new PLP formalism, called Bayesian knowledge bases. A Bayesian knowledge base 
consists mainly of a logic program that defines a direct influence relation
over a space of random variables. In Section 4,
we establish a declarative semantics for a Bayesian knowledge base 
based on a key notion of influence clauses. 
Influence clauses contain only ground atoms from the 
space of random variables and define the same direct influence relation
as the original Bayesian knowledge base does. In Section 5,
we present algorithms for
building a two-slice dynamic Bayesian network 
from a Bayesian knowledge base. We describe related work in Section 6 and 
summarize our work in Section 7.

\section{Preliminaries and Notation}
\label{subsec-1-1}
We assume the reader is familiar with basic ideas of
Bayesian networks \cite{pearl88} and logic programming \cite{Ld87}.
In particular, we assume the reader is familiar with  
the well-founded semantics \cite{VRS91} as well as 
SLG-resolution \cite{CSW95}.
Here we review some basic concepts concerning dynamic
Bayesian networks (DBNs). DBNs are introduced to
model the evolution of the state of the 
environment over time \cite{KKR95}. Briefly, a DBN is a Bayesian network whose random
variables are subscripted with time steps (basic units of time)
or time slices (i.e. intervals). 
In this paper, we use time slices. For instance, 
$Weather_{t-1}$, $Weather_t$ and $Weather_{t+1}$ are random 
variables representing the weather situations in time slices $t-1$, $t$ and
$t+1$, respectively. We can then use a DBN to 
depict how $Weather_{t-1}$ influences $Weather_t$.

\begin{figure}[htb]
\begin{center}
\setlength{\unitlength}{3079sp}%
\begingroup\makeatletter\ifx\SetFigFont\undefined%
\gdef\SetFigFont#1#2#3#4#5{%
  \reset@font\fontsize{#1}{#2pt}%
  \fontfamily{#3}\fontseries{#4}\fontshape{#5}%
  \selectfont}%
\fi\endgroup%
\begin{picture}(5394,615)(1829,-436)
\put(5851, 14){\makebox(0,0)[lb]{\smash{\SetFigFont{9}{10.8}{\rmdefault}{\mddefault}{\updefault}{\color[rgb]{0,0,0}$D_t$}%
}}}
\put(5851,-436){\makebox(0,0)[lb]{\smash{\SetFigFont{9}{10.8}{\familydefault}{\mddefault}{\updefault}{\color[rgb]{0,0,0}$B_t$}%
}}}
\put(5251,-436){\makebox(0,0)[lb]{\smash{\SetFigFont{9}{10.8}{\familydefault}{\mddefault}{\updefault}{\color[rgb]{0,0,0}$C_t$}%
}}}
\put(6451,-436){\makebox(0,0)[lb]{\smash{\SetFigFont{9}{10.8}{\rmdefault}{\mddefault}{\updefault}{\color[rgb]{0,0,0}$A_t$}%
}}}
\put(3226, 14){\makebox(0,0)[lb]{\smash{\SetFigFont{9}{10.8}{\rmdefault}{\mddefault}{\updefault}{\color[rgb]{0,0,0}$D_{t-1}$}%
}}}
\put(2401,-436){\makebox(0,0)[lb]{\smash{\SetFigFont{9}{10.8}{\familydefault}{\mddefault}{\updefault}{\color[rgb]{0,0,0}$C_{t-1}$}%
}}}
\put(4051,-436){\makebox(0,0)[lb]{\smash{\SetFigFont{9}{10.8}{\rmdefault}{\mddefault}{\updefault}{\color[rgb]{0,0,0}$A_{t-1}$}%
}}}
\put(3226,-436){\makebox(0,0)[lb]{\smash{\SetFigFont{9}{10.8}{\familydefault}{\mddefault}{\updefault}{\color[rgb]{0,0,0}$B_{t-1}$}%
}}}
\thinlines
{\color[rgb]{0,0,0}\put(5926,-61){\vector( 0,-1){200}}
}%
{\color[rgb]{0,0,0}\put(5476,-361){\vector( 1, 0){300}}
}%
{\color[rgb]{0,0,0}\put(6091,-361){\vector( 1, 0){300}}
}%
\thicklines
{\color[rgb]{0,0,0}\multiput(6801,-361)(66.66667,0.00000){7}{\makebox(8.5470,12.8205){\SetFigFont{10}{12}{\rmdefault}{\mddefault}{\updefault}.}}
}%
\thinlines
{\color[rgb]{0,0,0}\put(3301,-61){\vector( 0,-1){200}}
}%
\thicklines
{\color[rgb]{0,0,0}\multiput(1851,-361)(66.66667,0.00000){7}{\makebox(8.5470,12.8205){\SetFigFont{10}{12}{\rmdefault}{\mddefault}{\updefault}.}}
}%
\thinlines
{\color[rgb]{0,0,0}\put(4576,-361){\vector( 1, 0){540}}
}%
{\color[rgb]{0,0,0}\put(3691,-361){\vector( 1, 0){300}}
}%
{\color[rgb]{0,0,0}\put(2851,-361){\vector( 1, 0){300}}
}%
\end{picture}
\end{center}
\caption{A stationary DBN structure.} 
\label{dbn-fig}
\end{figure}
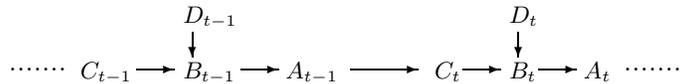

A DBN is represented by describing the intra-probabilistic relations
between random variables in each individual time slice $t$ ($t>0$) and
the inter-probabilistic relations between the random variables of each
two consecutive time slices $t-1$ and $t$. 
If both the intra- and inter-probabilistic relations
are the same for all time slices (in this case, the DBN is a repetition
of a Bayesian network over time; see Figure \ref{dbn-fig}), 
the DBN is called a {\em stationary} DBN  \cite{RN95};
otherwise it is called a {\em flexible} DBN \cite{glesner95}.
As far as we know, most existing DBN systems 
reported in the literature are stationary DBNs.

In a stationary DBN as shown in Figure \ref{dbn-fig}, 
the state evolution is determined by random variables like
$C$, $B$ and $A$, as they appear periodically and influence one another over time
(i.e., they produce cycles of direct influences). 
Such variables are called {\em state variables}. Note that $D$ is not a state variable.
Due to the characteristic of stationarity, a stationary DBN is often compactly represented 
as a two-slice DBN.

\begin{definition}
{\em
A {\em two-slice} DBN for a stationary DBN consists of two consecutive time slices,
$t-1$ and $t$, which describes (1) the intra-probabilistic relations
between the random variables in slice $t$ and (2)
the inter-probabilistic relations between the random variables 
in slice $t-1$ and the random variables in slice $t$.
}
\end{definition}

A two-slice DBN models a feedback system, where a cycle of direct influences
establishes a feedback connection. For convenience,
we depict feedback connections with dashed edges. Moreover, we refer to
nodes coming from slice $t-1$ as {\em state input nodes} (or {\em state input 
variables}).\footnote{When no confusion would occur, we will refer to nodes and 
random variables exchangeably.} 

\begin{example}
{\em
The stationary DBN of Figure \ref{dbn-fig} can be represented
by a two-slice DBN as shown in Figure \ref{feedback},
where $A$, $C$ and $B$ form a cycle of direct influences
and thus establish a feedback connection. 
This stationary DBN can also be represented
by a two-slice DBN starting from a different state input node 
such as $C_{t-1}$ or $B_{t-1}$.
These two-slice DBN structures are equivalent in the sense that
they model the same cycle of direct influences and 
can be unrolled into the same stationary DBN (Figure \ref{dbn-fig}).
}
\end{example}

\begin{figure}
\begin{minipage}[b]{0.4\linewidth} 
\centering
\setlength{\unitlength}{3079sp}%
\begingroup\makeatletter\ifx\SetFigFont\undefined%
\gdef\SetFigFont#1#2#3#4#5{%
  \reset@font\fontsize{#1}{#2pt}%
  \fontfamily{#3}\fontseries{#4}\fontshape{#5}%
  \selectfont}%
\fi\endgroup%
\begin{picture}(2924,1018)(839,-773)
\thinlines
{\color[rgb]{0,0,0}\put(851,-286){\line( 1, 0){900}}
}%
{\color[rgb]{0,0,0}\put(1726,-286){\vector( 1, 0){200}}
}%
{\color[rgb]{0,0,0}\put(1651,-526){\framebox(1800,759){}}
}%
{\color[rgb]{0,0,0}\put(2626,-11){\vector( 0,-1){200}}
}%
{\color[rgb]{0,0,0}\put(2176,-286){\vector( 1, 0){300}}
}%
{\color[rgb]{0,0,0}\put(2776,-286){\vector( 1, 0){300}}
}%
{\color[rgb]{0,0,0}\multiput(3386,-286)(146.00000,0.00000){3}{\line( 1, 0){ 73.000}}
}%
{\color[rgb]{0,0,0}\multiput(3751,-286)(0.00000,-135.71429){4}{\line( 0,-1){ 67.857}}
\multiput(3751,-761)(-154.54545,0.00000){17}{\line(-1, 0){ 77.273}}
\multiput(1201,-761)(0.00000,135.14286){4}{\line( 0, 1){ 67.571}}
\put(1201,-288){\vector( 0, 1){0}}
}%
\put(3151,-361){\makebox(0,0)[lb]{\smash{\SetFigFont{9}{10.8}{\rmdefault}{\mddefault}{\updefault}{\color[rgb]{0,0,0}$A_t$}%
}}}
\put(1126,-211){\makebox(0,0)[lb]{\smash{\SetFigFont{9}{10.8}{\rmdefault}{\mddefault}{\updefault}{\color[rgb]{0,0,0}$A_{t-1}$}%
}}}
\put(2551, 14){\makebox(0,0)[lb]{\smash{\SetFigFont{9}{10.8}{\rmdefault}{\mddefault}{\updefault}{\color[rgb]{0,0,0}$D_t$}%
}}}
\put(1951,-361){\makebox(0,0)[lb]{\smash{\SetFigFont{9}{10.8}{\familydefault}{\mddefault}{\updefault}{\color[rgb]{0,0,0}$C_t$}%
}}}
\put(2551,-361){\makebox(0,0)[lb]{\smash{\SetFigFont{9}{10.8}{\familydefault}{\mddefault}{\updefault}{\color[rgb]{0,0,0}$B_t$}%
}}}
\end{picture}
\caption{A two-slice DBN structure (a feedback system).} 
\label{feedback}
\end{minipage}
\hspace{0.9cm} 
\begin{minipage}[b]{0.4\linewidth}
\centering
\setlength{\unitlength}{3079sp}%
\begingroup\makeatletter\ifx\SetFigFont\undefined%
\gdef\SetFigFont#1#2#3#4#5{%
  \reset@font\fontsize{#1}{#2pt}%
  \fontfamily{#3}\fontseries{#4}\fontshape{#5}%
  \selectfont}%
\fi\endgroup%
\begin{picture}(2924,1018)(839,-773)
\thinlines
{\color[rgb]{0,0,0}\put(851,-286){\line( 1, 0){900}}
}%
{\color[rgb]{0,0,0}\put(1726,-286){\vector( 1, 0){200}}
}%
{\color[rgb]{0,0,0}\put(1651,-526){\framebox(1800,759){}}
}%
{\color[rgb]{0,0,0}\put(2626,-11){\vector( 0,-1){200}}
}%
{\color[rgb]{0,0,0}\put(2176,-286){\vector( 1, 0){300}}
}%
{\color[rgb]{0,0,0}\put(2776,-286){\vector( 1, 0){300}}
}%
{\color[rgb]{0,0,0}\multiput(3386,-286)(146.00000,0.00000){3}{\line( 1, 0){ 73.000}}
}%
{\color[rgb]{0,0,0}\multiput(3751,-286)(0.00000,-135.71429){4}{\line( 0,-1){ 67.857}}
\multiput(3751,-761)(-154.54545,0.00000){17}{\line(-1, 0){ 77.273}}
\multiput(1201,-761)(0.00000,135.14286){4}{\line( 0, 1){ 67.571}}
\put(1201,-288){\vector( 0, 1){0}}
}%
\put(3151,-361){\makebox(0,0)[lb]{\smash{\SetFigFont{9}{10.8}{\rmdefault}{\mddefault}{\updefault}{\color[rgb]{0,0,0}$A$}%
}}}
\put(1126,-211){\makebox(0,0)[lb]{\smash{\SetFigFont{9}{10.8}{\rmdefault}{\mddefault}{\updefault}{\color[rgb]{0,0,0}$A_{t-1}$}%
}}}
\put(2551, 14){\makebox(0,0)[lb]{\smash{\SetFigFont{9}{10.8}{\rmdefault}{\mddefault}{\updefault}{\color[rgb]{0,0,0}$D$}%
}}}
\put(1951,-361){\makebox(0,0)[lb]{\smash{\SetFigFont{9}{10.8}{\familydefault}{\mddefault}{\updefault}{\color[rgb]{0,0,0}$C$}%
}}}
\put(2551,-361){\makebox(0,0)[lb]{\smash{\SetFigFont{9}{10.8}{\familydefault}{\mddefault}{\updefault}{\color[rgb]{0,0,0}$B$}%
}}}
\end{picture}
\caption{A simplified two-slice DBN structure.} 
\label{feedback2}
\end{minipage}
\end{figure}

Observe that in a two-slice DBN, all random variables except state input nodes
have the same subscript $t$. In the sequel, the subscript $t$ is 
omitted for simplification of the structure.
For instance, the two-slice DBN of Figure \ref{feedback} 
is simplified to that of Figure \ref{feedback2}.

In the rest of this section, we introduce some necessary notation for logic programs. 
Variables begin with a capital
letter, and predicate, function and constant symbols with a lower-case letter.
We use $p(.)$ to refer to any predicate/atom whose predicate symbol is $p$ and use
$p(\overrightarrow{X})$ to refer to $p(X_1,...,X_n)$ where all $X_i$s are variables.
There is one special predicate, $true$, which is always logically true.
A predicate $p(\overrightarrow{X})$ is {\em typed}
if its arguments $\overrightarrow{X}$
are typed so that each argument takes 
on values in a well-defined finite domain. 
A (general) logic program $P$ is a finite set of clauses of the form
\begin{equation}
\label{eq1}
A\leftarrow B_1,...,B_m, \neg  C_1, ..., \neg  C_n
\end{equation}
where $A$, the $B_i$s and $C_j$s are atoms.
We use $HU(P)$ and $HB(P)$ to denote the Herbrand universe and Herbrand base of $P$, respectively,
and use $WF(P) = <$$I_t, I_f$$>$ to denote
the well-founded model of $P$, where
$I_t, I_f \subseteq HB(P)$, and every $A$ in $I_t$ is true
and every $A$ in $I_f$ is false in $WF(P)$.
By a {\em (Herbrand) ground instance} of a clause/atom $C$ we refer to a ground
instance of $C$ that is obtained by replacing
all variables in $C$ with some terms in $HU(P)$.

A logic program $P$ is a {\em positive} logic program
if no negative literal occurs in the body of any clause. $P$ is a {\em Datalog} program if
no clause in $P$ contains function symbols.
$P$ is an {\em acyclic} logic program if there is a mapping
$map$ from the set of ground instances of atoms in $P$ into the set of natural numbers
such that for any ground instance $A\leftarrow B_1,...,B_k, \neg B_{k+1}, ..., \neg B_n$
of any clause in $P$, $map(A) > map(B_i)$ $(1\leq i\leq n)$ \cite{apt91}.
$P$ is said to have the {\em bounded-term-size property} 
w.r.t. a set of predicates $\{p_1(.), ..., p_t(.)\}$ if there is a function
$f(n)$ such that for any $1\leq i\leq t$
whenever a top goal $G_0 = \leftarrow p_i(.)$ has no argument whose term size exceeds
$n$, no atoms in any SLDNF- (or SLG-) derivations for $G_0$
have an argument whose term size exceeds $f(n)$ (this definition
is adapted from \cite{VG89}).

\section{Definition of a Bayesian Knowledge Base}
In this section, we introduce a new PLP formalism,
called Bayesian knowledge bases. Bayesian knowledge bases
accommodate recursive loops and define the direct influence 
relation in terms of the well-founded semantics.

\begin{definition}
\label{kb}
{\em
A {\em Bayesian knowledge base} is a triple $<$$PB\cup CB,T_x,CR$$>$, where 
\begin{itemize}
\item 
$PB\cup CB$ is a logic program,
each clause in $PB$ being of the form
\begin{eqnarray}
\label{cll-2}
p(.) \leftarrow \underbrace{p_1(.),...,p_l(.)}_{direct \ influences}, true, 
\underbrace{B_1,...,B_m, \neg  C_1, ..., \neg  C_n}_{context}, \nonumber \\
  \hspace{.1in}  
\underbrace{member(X_1,DOM_1),...,member(X_s,DOM_s)}_{type \ constraints}
\end{eqnarray}
where (i) the predicate symbols $p,p_1,...,p_l$ only occur in $PB$ and 
(ii) $p(.)$ is typed so that for each variable $X_i$ in it with 
a finite domain $DOM_i$ (a list of constants)
there is an atom $member(X_i,DOM_i)$ in the clause body.
\item
$T_x$ is a set of conditional probability 
tables (CPTs) of the form ${\bf P}(p(.)|p_1(.),...,$
$p_l(.))$, each being attached to 
a clause (\ref{cll-2}) in $PB$.
\item
$CR$ is a combination rule such as {\em noisy-or, min} or {\em max} \cite{Kersting2000,ngo-had97,RN95}.
\end{itemize}
}
\end{definition}

A Bayesian knowledge base contains a logic program
that can be divided into two parts, $PB$ and $CB$.
$PB$ defines a direct influence relation, each clause
(\ref{cll-2}) saying that the atoms $p_1(.), ..., p_l(.)$ 
have direct influences on $p(.)$ in the context that
$B_1,...,B_m, \neg  C_1, ..., \neg  C_n,$ $member(X_1,DOM_1),...,$ $member(X_s,DOM_s)$ 
is true in $PB\cup CB$
under the well-founded semantics. Note that the special literal
$true$ is used in clause (\ref{cll-2}) to mark the beginning
of the context; it is always true in the well-founded model $WF(PB\cup CB)$.
For each variable $X_i$ in the head $p(.)$, $member(X_i,DOM_i)$ is used
to enforce the type constraint on $X_i$, i.e. the value of $X_i$ comes from its domain
$DOM_i$. $CB$ assists $PB$ in defining the direct influence relation
by introducing some auxiliary predicates (such as $member(.)$) to describe
contexts.\footnote{The predicate $true$ can be defined
in $CB$ using a unit clause.} Clauses in $CB$ do not describe direct
influences. 

Recursive loops are allowed in $PB$ and $CB$.
In particular, when some $p_i(.)$ in clause (\ref{cll-2}) is
the same as the head $p(.)$, a cyclic direct influence occurs.
Such a cyclic influence models a feedback connection
and is interpreted as $p(.)$ at present depending
on itself in the past. 

In this paper, we focus on Datalog programs,
although the proposed approach applies to logic programs with 
the bounded-term-size property (w.r.t. the set of predicates 
appearing in the heads of clauses in $PB$) as well.
Datalog programs are widely used in database and knowledge base systems \cite{ullman88}
and have a polynomial time complexity
in computing their well-founded models \cite{VRS91}.
In the sequel, we assume that except for the predicate
$member(.)$, $PB\cup CB$ is a Datalog program. 

For each clause (\ref{cll-2}) in $PB$, there is a unique
CPT, ${\bf P}(p(.)|p_1(.),...,p_l(.))$, in $T_x$
specifying the degree of the direct influences. Such a CPT
is shared by all instances of clause (\ref{cll-2}). 

A Bayesian knowledge base has the following important property.

\begin{theorem}
\label{th-ground}
(1) All unit clauses in $PB$ are ground. 
(2) Let $G_0=\leftarrow p(.)$ be a goal with
$p$ being a predicate symbol occurring in the head of a clause in $PB$.
Then all answers of $G_0$ derived from $PB\cup CB\cup\{G_0\}$
by applying SLG-resolution are ground.
\end{theorem}

\noindent {\bf Proof:} 
(1) If the head of a clause in $PB$
contains variables, there must be atoms of the form $member(X_i,DOM_i)$
in its body. This means that clauses whose head contains variables are not unit clauses.
Therefore, all unit clauses in $PB$ are ground.

(2) Let $A$ be an answer of $G_0$ obtained by applying SLG-resolution 
to $PB\cup CB\cup\{G_0\}$. Then $A$ must be produced
by applying a clause in $PB$ of form (\ref{cll-2})
with a most general unifier (mgu) $\theta$ such that $A=p(.)\theta$ and the body 
$(p_1(.), ..., p_l(.), true, B_1,...,B_m, \neg  C_1, ..., \neg  C_n,$ 
$member(X_1,DOM_1), ...,member($ $X_s,DOM_s))\theta$ is evaluated
true in the well-founded model $WF(PB\cup CB)$. Note that the type constraints
$(member(X_1,DOM_1), ..., member(X_s,DOM_s))\theta$ being 
evaluated true by SLG-resolution guarantees
that all variables $X_i$s in the head $p(.)$ are instantiated by $\theta$ into
constants in their domains $DOM_i$s. This means that $A$ is ground. $\Box$\\[.1in]
\indent
For the sake of simplicity, in the sequel for each clause 
(\ref{cll-2}) in $PB$, we omit its type constraints
$member(X_i,DOM_i)$ ($1\leq i\leq s$). Therefore, when we say that the context
$B_1,...,B_m,$ $\neg C_1,..., \neg C_n$ is true, we assume that
the related type constraints are true as well.

\begin{example}
\label{aids-eg}
{\em
We borrow the well-known AIDS program from \cite{glesner95} 
(a simplified version) as a running
example to illustrate our PLP approach. It is formulated by
a Bayesian knowledge base $KB_1$ with the following logic program:\footnote{This Bayesian 
knowledge base $KB_1=<$$PB_1\cup CB_1,T_{x_1},CR_1$$>$ 
may well contain contexts that describe a person's background
information. The contexts together with $CB_1$, $T_{x_1}$ and $CR_1$ 
are omitted here for the sake of simplicity.}
\begin{tabbing}
$\qquad\qquad PB_1:\ $ \= 1. $aids(p1).$\\
\> 2. $aids(p3).$\\
\> 3. $aids(X)\leftarrow aids(X).$\\
\> 4. $aids(X)\leftarrow aids(Y), contact(X,Y).$\\
\> 5. $contact(p1, p2).$\\
\> 6. $contact(p2, p1).$
\end{tabbing}
Note that both the 3rd and the 4-th clause produce recursive loops.
The 3rd clause also has a cyclic direct influence.
Conceptually, the two clauses model the fact that the direct
influences on $aids(X)$ come from whether $X$ was infected with 
aids earlier (the feedback connection induced from the 3rd clause) or whether
$X$ has contact with someone $Y$ who is infected with aids (the 4-th clause).
}
\end{example}

\section{Declarative Semantics}
In this section, we formally describe the space of random variables and 
the direct influence relation defined by a Bayesian knowledge base $KB$. 
We then define probability distributions induced by $KB$.

\subsection{Space of Random Variables and Influence Clauses}
\label{sec-build-bn}

A Bayesian knowledge base $KB$
defines a direct influence relation over a subset of $HB(PB)$.  
Recall that any random variable in a Bayesian network 
is either an input node (with no parent nodes) or
a node on which some other nodes (i.e. its parent nodes) 
in the network have direct influences.
Since an input node can be viewed as a node whose direct influences
come from an empty set of parent nodes, we can define a space of random variables 
from a Bayesian knowledge base $KB$
by taking all unit clauses in $PB$ as input nodes and deriving the 
other nodes iteratively based on the direct influence relation 
defined by $PB$. Formally, we have

\begin{definition}
\label{bn-node}
{\em
The {\em space of random variables} of $KB$, denoted ${\cal S}(KB)$, 
is recursively defined as follows:
\begin{enumerate}
\item
All unit clauses in $PB$ are random variables in ${\cal S}(KB)$.
\item
Let 
$A\leftarrow A_1,...,A_l, true, B_1,...,B_m, \neg  C_1, ..., \neg  C_n$ 
be a ground instance of a clause in $PB$.
If the context $B_1,...,B_m, \neg C_1,..., \neg C_n$
is true in the well-founded  model $WF(PB\cup CB)$ and 
$\{A_1,...,A_l\}$ $\subseteq {\cal S}(KB)$, then $A$ is a random variable in ${\cal S}(KB)$. In
this case, each $A_i$ is said to have a {\em direct influence} on $A$.
\item
${\cal S}(KB)$ contains only those ground atoms satisfying the above two conditions.
\end{enumerate}
}
\end{definition}

\begin{definition}
\label{inf-by}
{\em
For any random variables $A$, $B$ in ${\cal S}(KB)$, we say $A$ is {\em influenced by} $B$ if
$B$ has a direct influence on $A$, or for some $C$ in ${\cal S}(KB)$ $A$ is influenced by $C$
and $C$ is influenced by $B$. A {\em cyclic influence} occurs if $A$ is influenced by itself.
}
\end{definition}

\begin{example}[Example \ref{aids-eg} continued]
\label{aids-eg-space}
{\em
The clauses 1, 2, 5 and 6 are unit clauses, thus random variables.
$aids(p2)$ is then derived applying the 4-th clause. Consequently,
${\cal S}(KB_1) = \{aids(p1), aids(p2), aids(p3),
contact(p1, p2), contact(p2, p1)\}$. 
$aids(p1)$ and $aids(p2)$ have a direct influence on each other.
There are three cyclic influences:
$aids(pi)$ is influenced by itself for each $i=1,2,3$. 
}  
\end{example}

Let $WF(PB\cup CB) = <$$I_t, I_f$$>$ be the well-founded model
of $PB\cup CB$ and let 
$I_{PB} = \{p(.)\in I_t | p$ occurs in the head of some clause in $PB\}$. 
The following result shows that the space of random variables
is uniquely determined by the well-founded model.

\begin{theorem}
\label{v-is-ipb}
${\cal S}(KB) = I_{PB}$.
\end{theorem}

\noindent {\bf Proof:} First note that all 
unit clauses in $PB$
are both in ${\cal S}(KB)$ and in $I_{PB}$.
We prove this theorem by induction on 
the maximum depth $d\geq 0$ of backward derivations
of a random variable $A$. 

($\Longrightarrow$) Let $A\in {\cal S}(KB)$. When $d=0$, $A$ is a 
unit clause in $PB$, so $A\in I_{PB}$. 
For the induction step, assume $B\in I_{PB}$ for any $B\in {\cal S}(KB)$ whose
maximum depth $d$ of backward derivations is below $k$.
Let $d=k$ for $A$. There must be a ground instance
$A \leftarrow A_1,...,A_l, true, B_1,...,B_m, \neg  C_1, ..., \neg  C_n$ 
of a clause in $PB$ such that the $A_i$s are already in ${\cal S}(KB)$ and
$B_1,...,B_m, \neg  C_1, ...,$ $\neg  C_n$ is
true in the well-founded model $WF(PB\cup CB)$. Since the head $A$
is derived from the $A_i$s in the body, the maximum depth
for each $A_i$ must be below the depth $k$ for the head $A$.
By the induction hypothesis, the $A_i$s are in $I_{PB}$. 
By definition of the well-founded model, 
$A$ is true in $WF(PB\cup CB)$ and thus $A\in I_{PB}$. 

($\Longleftarrow$) Let $A\in I_{PB}$. When $d=0$, $A$ is a 
unit clause
in $PB$, so $A\in {\cal S}(KB)$. 
For the induction step, assume $B\in {\cal S}(KB)$ for any $B\in I_{PB}$ whose
maximum depth $d$ of backward derivations is below $k$.
Let $d=k$ for $A$. There must be a ground instance
$A \leftarrow A_1,...,A_l, true, ...$ 
of a clause in $PB$ such that the body is true in $WF(PB\cup CB)$. 
Note that the predicate symbol of each $A_i$
occurs in the head of a clause in $PB$. Since the head $A$ is derived
from the literals in the body, the maximum depth of backward derivations
for each $A_i$ in the body must be below the depth $k$ for the head $A$.
By the induction hypothesis, the $A_i$s are in ${\cal S}(KB)$. 
By Definition \ref{bn-node}, $A\in {\cal S}(KB)$.  $\Box$\\[.1in]
\indent 
Theorem \ref{v-is-ipb} suggests that the space of random variables 
can be computed by applying an existing procedure for the well-founded 
model such as SLG-resolution or SLTNF-resolution. 
Since SLG-resolution has been implemented 
as the well-known $XSB$ system \cite{SSWFR98},
in this paper we apply it for the PLP backward-chaining inferences.
SLG-resolution is a tabling
mechanism for top-down computation of the well-founded model. 
For any atom $A$, during the process of 
evaluating a goal $\leftarrow A$, SLG-resolution
stores all answers of $A$ 
in a space called {\em table}, denoted ${\cal T}_A$. 

Let $\{p_1,...,p_t\}$ be the set of 
predicate symbols occurring in the heads of clauses in $PB$,
and let $GS_0 = \{\leftarrow p_1(\overrightarrow{X_1}), ..., \leftarrow 
p_t(\overrightarrow{X_t})\}$.\\[.1in]
{\bf Algorithm 1: Computing random variables.}
\begin{enumerate}
\item
${\cal S'}(KB) = \emptyset$.

\item
For each $\leftarrow p_i(\overrightarrow{X_i})$ in $GS_0$
\begin{enumerate}
\item
Compute the goal $\leftarrow p_i(\overrightarrow{X_i})$ 
by applying SLG-resolution to $PB\cup CB \cup \{\leftarrow p_i(\overrightarrow{X_i})\}$.

\item
${\cal S'}(KB) = {\cal S'}(KB) \cup {\cal T}_{p_i(\overrightarrow{X_i})}$.
\end{enumerate}

\item
Return ${\cal S'}(KB)$.
\end{enumerate}
\begin{theorem}
\label{th-vi}
Algorithm 1 terminates, yielding a finite set ${\cal S'}(KB) = {\cal S}(KB)$.
\end{theorem}

\noindent {\bf Proof:} Let $WF(PB\cup CB) = <$$I_t, I_f$$>$ be the well-founded model
of $PB\cup CB$. By the soundness and completeness of 
SLG-resolution, Algorithm 1 will terminate
with a finite output ${\cal S'}(KB)$ that consists of all answers of 
$p_i(\overrightarrow{X_i})$ ($1\leq i\leq t$). By Theorem \ref{th-ground},
all answers in ${\cal S'}(KB)$ are ground. This means ${\cal S'}(KB) = I_{PB}$.
Hence, by Theorem \ref{v-is-ipb} ${\cal S'}(KB) = {\cal S}(KB)$. $\Box$\\[.1in]
\indent We introduce the following principal concept.

\begin{definition}
\label{inf-clause}
{\em
Let $A \leftarrow A_1,...,A_l, true, B_1,...,B_m, \neg  C_1, ..., \neg  C_n$
be a ground instance of the $k$-th clause in $PB$ such that its body
is true in the well-founded model $WF(PB\cup CB)$. We call
\begin{equation}
\label{infcl}
k. \ \ A \leftarrow A_1,...,A_l
\end{equation}
an {\em influence clause}.\footnote{The prefix ``$k.$'' would be omitted sometimes
for the sake of simplicity.} All influence clauses derived from 
all clauses in $PB$ constitute the {\em set of influence clauses} of $KB$, 
denoted ${\cal I}_{clause}(KB)$.  
}
\end{definition}

The following result is immediate from Definition \ref{bn-node}
and Theorem \ref{v-is-ipb}.

\begin{theorem}
\label{icl-ground}
For any influence clause (\ref{infcl}),
$A$ and all $A_i$s are random variables in ${\cal S}(KB)$. 
\end{theorem}

Influence clauses have the following principal property.

\begin{theorem}
\label{main-inf}
For any $A_i$ and $A$ in HB(PB),
$A_i$ has a direct influence on $A$, which
is derived from the $k$-th clause in $PB$,
if and only if there is an influence clause in ${\cal I}_{clause}(KB)$
of the form $k. \ A \leftarrow A_1,...,A_i, ..., A_l$.
\end{theorem}

\noindent {\bf Proof:} 
($\Longrightarrow$)  Assume $A_i$ has a direct influence on $A$, which
is derived from the $k$-th clause in $PB$. By Definition \ref{bn-node},
the $k$-th clause has a ground instance of the form
$A\leftarrow A_1,...,A_i,...,A_l, true, B_1,...,B_m, \neg  C_1, ..., \neg  C_n$
such that $B_1,...,B_m,$ $\neg C_1,..., \neg C_n$
is true in $WF(PB\cup CB)$ and $\{A_1,...,A_i,...,A_l\}$ $\subseteq {\cal S}(KB)$.
By Theorem \ref{v-is-ipb}, $A_1,...,A_i,...,A_l$ is true in $WF(PB\cup CB)$.
Thus, $k. \ A \leftarrow A_1,...,A_i, ..., A_l$ is an 
influence clause in ${\cal I}_{clause}(KB)$.

($\Longleftarrow$) Assume that ${\cal I}_{clause}(KB)$ contains
an influence clause
$k. \ A \leftarrow A_1,...,$ $A_i, ..., A_l$. Then the $k$-th clause in $PB$
has a ground instance of the form
$A\leftarrow A_1,...,A_i,..., A_l, true, B_1,$ $...,B_m, \neg  C_1, ..., \neg  C_n$ 
such that its body
is true in $WF($ $PB\cup CB)$ and (by Theorem \ref{icl-ground}) 
$\{A_1,...,A_i,...,A_l\}\subseteq {\cal S}(KB)$.
By Definition \ref{bn-node}, 
$A\in {\cal S}(KB)$ and $A_i$ has a direct influence on $A$. $\Box$\\[.1in]
\indent
The following result is immediate from Theorem \ref{main-inf}.

\begin{corollary}
\label{corr1}
For any atom $A$, $A$ is in ${\cal S}(KB)$ 
if and only if there is an influence clause in ${\cal I}_{clause}(KB)$
whose head is $A$.
\end{corollary} 

Theorem \ref{main-inf} shows the significance of influence clauses:
they define the same direct influence relation over the same space
of random variables as the original Bayesian knowledge base does.
Therefore, a Bayesian network can be built directly from ${\cal I}_{clause}(KB)$ 
provided the influence clauses are available. 

Observe that to compute the space of random variables (see Algorithm 1), 
SLG-resolution will construct a proof tree
rooted at the goal $\leftarrow p_i(\overrightarrow{X_i})$ for each $1\leq i\leq t$ 
\cite{CSW95}. For each answer
$A$ of $p_i(\overrightarrow{X_i})$ in ${\cal S}(KB)$ there must be a success
branch (i.e. a branch starting at the root node
and ending at a node marked with {\em success}) in the tree
that generates the answer. Let
$p_i(.) \leftarrow A_1,...,A_l, true, ...$
be the $k$-th clause in $PB$ that is applied to expand the root goal
$\leftarrow p_i(\overrightarrow{X_i})$ in the success branch 
and let $\theta$ be the composition of all mgus along the branch.
Then $A = p_i(.)\theta$ and the body $A_1,...,A_l, true, ...$ 
is evaluated true, with the mgu $\theta$, in $WF(PB\cup CB)$ 
by SLG-resolution. This means that for each $1\leq j\leq l$,   
$A_j\theta$ is an answer of $A_j$ that is  
derived by applying SLG-resolution to $PB\cup CB\cup \{\leftarrow A_j'\}$ 
where $A_j'$ is $A_j$ or some instance of $A_j$. By Theorem \ref{th-ground}, 
all $A_j\theta$s are ground atoms. Therefore,
$k. \ p_i(.)\theta \leftarrow A_1\theta,...,A_l\theta$ 
is an influence clause. Hence we have the following result.

\begin{theorem}
\label{inf-branch}
Let $B_r$ be a success branch in a proof tree of SLG-resolution, 
$p_i(.) \leftarrow A_1,...,A_l,$ $true, ...$ be the $k$-th
clause in $PB$ that expands the root goal in $B_r$,
and $\theta$ be the composition of all mgus along $B_r$.
$B_r$ produces an influence clause
$k. \ p_i(.)\theta \leftarrow A_1\theta,...,A_l\theta$.
\end{theorem}

Every success branch in a proof tree for a goal in $GS_0$ 
produces an influence clause.
The set of influence clauses can then be obtained by collecting
all influence clauses from all such proof trees in SLG-resolution.\\[.1in]
{\bf Algorithm 2: Computing influence clauses.}
\begin{enumerate}
\item
For each goal $\leftarrow p_i(\overrightarrow{X_i})$ in $GS_0$,
compute all answers of $p_i(\overrightarrow{X_i})$ 
by applying SLG-resolution to $PB\cup CB \cup \{\leftarrow p_i(\overrightarrow{X_i})\}$ 
while for each success branch starting 
at the root goal $\leftarrow p_i(\overrightarrow{X_i})$, 
collecting an influence clause from the branch into ${\cal I}_{clause}'(KB)$.

\item
Return ${\cal I}_{clause}'(KB)$.
\end{enumerate}

\begin{theorem}
\label{th-infcl}
Algorithm 2 terminates, yielding a finite set 
${\cal I}_{clause}'(KB) = {\cal I}_{clause}(KB)$.
\end{theorem}

\noindent {\bf Proof:} That Algorithm 2 terminates 
is immediate from Theorem \ref{th-vi}, as 
except for collecting influence clauses,
Algorithm 2 makes the same derivations as Algorithm 1.
The termination of Algorithm 2 then implies
${\cal I}_{clause}'(KB)$ is finite. 

By Theorem \ref{inf-branch}, any clause in ${\cal I}_{clause}'(KB)$
is an influence clause in ${\cal I}_{clause}(KB)$. We now prove
the converse. Let $k. \ A \leftarrow A_1, ..., A_l$
be an influence clause in ${\cal I}_{clause}(KB)$. 
Then the $k$-th clause in $PB$
$A'\leftarrow A_1',...,A_l', true, ...$. 
has a ground instance of the form
$A\leftarrow A_1,..., A_l, true, ...$ whose body
is true in $WF(PB\cup CB)$. By the completeness of SLG-resolution,
there must be a success branch
in the proof tree rooted at a goal $\leftarrow p_i(\overrightarrow{X_i})$ in $GS_0$
where (1) the root goal is expanded by the $k$-th clause, (2)
the composition of all mgus along the branch is $\theta$, 
and (3) $A\leftarrow A_1,...,A_l, true, ...$ is an instance of 
$(A'\leftarrow A_1',...,A_l', true, ...)\theta$.
By Theorem \ref{inf-branch}, $k. \ A'\theta\leftarrow A_1'\theta,...,A_l'\theta$ 
is an influence clause. Since any influence clause is ground, 
$k. \ A'\theta\leftarrow A_1'\theta,...,A_l'\theta$ is the same as
$k. \ A \leftarrow A_1, ..., A_l$. This influence clause from the success
branch will be collected into ${\cal I}_{clause}'(KB)$ by Algorithm 2. 
Thus, any clause in ${\cal I}_{clause}(KB)$
is in ${\cal I}_{clause}'(KB)$. $\Box$

\begin{example}[Example \ref{aids-eg-space} continued]
\label{aids-eg-infcl}
{\em
There are two predicate symbols, $aids$ and $contact$, in the heads of clauses in $PB_1$.
Let $GS_0 = \{\leftarrow aids(X), \leftarrow contact(Y, Z)\}$.
Algorithm 2 will generate two proof trees 
rooted at $\leftarrow aids(X)$ and $\leftarrow contact(Y, Z)$,
respectively, as shown in Figures \ref{slg-aids} and  \ref{slg-aids2}. 
In the proof trees, a label ${\cal C}_i$ on an edge 
indicates that the $i$-th clause in $PB$ 
is applied, and the other labels like $X=p1$
on an edge show that an answer from a table is applied.
Each success branch yields an influence clause. For instance, 
expanding the root goal $\leftarrow aids(X)$ by the 3rd clause
produces a child node $\leftarrow aids(X)$ (Figure \ref{slg-aids}).
Then applying the answers of $aids(X)$ from the table
${\cal T}_{aids(X)}$ to the goal of this node leads to
three success branches. Applying the mgu $\theta$ on each 
success branch to the 3rd clause yields three influence
clauses of the form 3. $aids(pi)\leftarrow aids(pi)$ 
($i=1,2,3$). As a result, we 
obtain the following set of influence clauses:
\begin{tabbing}
$\qquad {\cal I}_{clause}(KB_1):$ \= 1. $aids(p1).$\\
\> 2. $aids(p3).$\\
\> 3. $aids(p1)\leftarrow aids(p1).$\\
\> 3. $aids(p2)\leftarrow aids(p2).$\\
\> 3. $aids(p3)\leftarrow aids(p3).$\\
\> 4. $aids(p2)\leftarrow aids(p1), contact(p2,p1).$\\
\> 4. $aids(p1)\leftarrow aids(p2), contact(p1,p2).$\\
\> 5. $contact(p1, p2).$\\
\> 6. $contact(p2, p1).$
\end{tabbing}
}
\end{example}

\begin{figure}[htb]
\setlength{\unitlength}{3000sp}%
\begingroup\makeatletter\ifx\SetFigFont\undefined%
\gdef\SetFigFont#1#2#3#4#5{%
  \reset@font\fontsize{#1}{#2pt}%
  \fontfamily{#3}\fontseries{#4}\fontshape{#5}%
  \selectfont}%
\fi\endgroup%
\begin{picture}(6792,2161)(901,-1667)
\put(5926,-661){\makebox(0,0)[lb]{\smash{\SetFigFont{5}{6.0}{\rmdefault}{\mddefault}{\updefault}{\color[rgb]{0,0,0}$Y=p2$}%
}}}
\put(5551,-211){\makebox(0,0)[lb]{\smash{\SetFigFont{6}{7.2}{\rmdefault}{\mddefault}{\updefault}{\color[rgb]{0,0,0}$\leftarrow aids(Y), contact(X,Y)$}%
}}}
\put(5926,-1111){\makebox(0,0)[lb]{\smash{\SetFigFont{5}{6.0}{\rmdefault}{\mddefault}{\updefault}{\color[rgb]{0,0,0}$X=p1$}%
}}}
\put(4726,-1411){\makebox(0,0)[lb]{\smash{\SetFigFont{7}{8.4}{\rmdefault}{\mddefault}{\updefault}{\color[rgb]{0,0,0}$success$}%
}}}
\put(6226,-1411){\makebox(0,0)[lb]{\smash{\SetFigFont{7}{8.4}{\rmdefault}{\mddefault}{\updefault}{\color[rgb]{0,0,0}$success$}%
}}}
\put(7426,-1411){\makebox(0,0)[lb]{\smash{\SetFigFont{7}{8.4}{\rmdefault}{\mddefault}{\updefault}{\color[rgb]{0,0,0}$failed$}%
}}}
\put(4501,-1636){\makebox(0,0)[lb]{\smash{\SetFigFont{5}{6.0}{\rmdefault}{\mddefault}{\updefault}{\color[rgb]{0,0,0}$\theta = \{Y/p1,X/p2\}$}%
}}}
\put(6001,-1636){\makebox(0,0)[lb]{\smash{\SetFigFont{5}{6.0}{\rmdefault}{\mddefault}{\updefault}{\color[rgb]{0,0,0}$\theta = \{Y/p2,X/p1\}$}%
}}}
\put(1051,-211){\makebox(0,0)[lb]{\smash{\SetFigFont{7}{8.4}{\rmdefault}{\mddefault}{\updefault}{\color[rgb]{0,0,0}$success$}%
}}}
\put(901,-436){\makebox(0,0)[lb]{\smash{\SetFigFont{5}{6.0}{\rmdefault}{\mddefault}{\updefault}{\color[rgb]{0,0,0}$\theta = \{X/p1\}$}%
}}}
\put(1876,-211){\makebox(0,0)[lb]{\smash{\SetFigFont{7}{8.4}{\rmdefault}{\mddefault}{\updefault}{\color[rgb]{0,0,0}$success$}%
}}}
\put(2401,-736){\makebox(0,0)[lb]{\smash{\SetFigFont{7}{8.4}{\rmdefault}{\mddefault}{\updefault}{\color[rgb]{0,0,0}$success$}%
}}}
\put(1951,-1036){\makebox(0,0)[lb]{\smash{\SetFigFont{5}{6.0}{\rmdefault}{\mddefault}{\updefault}{\color[rgb]{0,0,0}$\theta = \{X/p1\}$}%
}}}
\put(5026,-1111){\makebox(0,0)[lb]{\smash{\SetFigFont{5}{6.0}{\rmdefault}{\mddefault}{\updefault}{\color[rgb]{0,0,0}$X=p2$}%
}}}
\put(1801,-436){\makebox(0,0)[lb]{\smash{\SetFigFont{5}{6.0}{\rmdefault}{\mddefault}{\updefault}{\color[rgb]{0,0,0}$\theta = \{X/p3\}$}%
}}}
\put(3751,-736){\makebox(0,0)[lb]{\smash{\SetFigFont{7}{8.4}{\rmdefault}{\mddefault}{\updefault}{\color[rgb]{0,0,0}$success$}%
}}}
\put(3676,-1036){\makebox(0,0)[lb]{\smash{\SetFigFont{5}{6.0}{\rmdefault}{\mddefault}{\updefault}{\color[rgb]{0,0,0}$\theta = \{X/p2\}$}%
}}}
\put(3076,-736){\makebox(0,0)[lb]{\smash{\SetFigFont{7}{8.4}{\rmdefault}{\mddefault}{\updefault}{\color[rgb]{0,0,0}$success$}%
}}}
\put(2851,-1036){\makebox(0,0)[lb]{\smash{\SetFigFont{5}{6.0}{\rmdefault}{\mddefault}{\updefault}{\color[rgb]{0,0,0}$\theta = \{X/p3\}$}%
}}}
\put(2251, 14){\makebox(0,0)[lb]{\smash{\SetFigFont{5}{6.0}{\rmdefault}{\mddefault}{\updefault}{\color[rgb]{0,0,0}${\cal C}_2$}%
}}}
\put(1501, 14){\makebox(0,0)[lb]{\smash{\SetFigFont{5}{6.0}{\rmdefault}{\mddefault}{\updefault}{\color[rgb]{0,0,0}${\cal C}_1$}%
}}}
\put(3451, 14){\makebox(0,0)[lb]{\smash{\SetFigFont{5}{6.0}{\rmdefault}{\mddefault}{\updefault}{\color[rgb]{0,0,0}${\cal C}_3$}%
}}}
\put(5026,-661){\makebox(0,0)[lb]{\smash{\SetFigFont{5}{6.0}{\rmdefault}{\mddefault}{\updefault}{\color[rgb]{0,0,0}$Y=p1$}%
}}}
\thinlines
{\color[rgb]{0,0,0}\put(3301,294){\line( 0,-1){130}}
}%
{\color[rgb]{0,0,0}\put(3301,164){\vector( 0,-1){225}}
}%
{\color[rgb]{0,0,0}\put(2776,-361){\vector( 0,-1){225}}
\put(2776,-361){\line( 1, 0){1050}}
\put(3826,-361){\vector( 0,-1){225}}
}%
{\color[rgb]{0,0,0}\put(3301,-231){\line( 0,-1){130}}
}%
{\color[rgb]{0,0,0}\put(6451,-436){\vector( 0,-1){300}}
}%
{\color[rgb]{0,0,0}\put(4951,-436){\vector( 0,-1){300}}
\put(4951,-436){\line( 1, 0){2700}}
\put(7651,-436){\vector( 0,-1){300}}
}%
{\color[rgb]{0,0,0}\put(4951,-961){\vector( 0,-1){300}}
}%
{\color[rgb]{0,0,0}\put(6451,-961){\vector( 0,-1){300}}
}%
{\color[rgb]{0,0,0}\put(7651,-961){\vector( 0,-1){300}}
}%
{\color[rgb]{0,0,0}\put(6451,-286){\line( 0,-1){130}}
}%
{\color[rgb]{0,0,0}\put(1401,164){\vector( 0,-1){225}}
\put(1401,164){\line( 1, 0){5050}}
\put(6451,164){\vector( 0,-1){225}}
}%
{\color[rgb]{0,0,0}\put(2101,164){\vector( 0,-1){225}}
}%
{\color[rgb]{0,0,0}\put(3301,-361){\vector( 0,-1){225}}
}%
\put(2851,389){\makebox(0,0)[lb]{\smash{\SetFigFont{6}{7.2}{\rmdefault}{\mddefault}{\updefault}{\color[rgb]{0,0,0}$\leftarrow aids(X)$}%
}}}
\put(2926,-211){\makebox(0,0)[lb]{\smash{\SetFigFont{6}{7.2}{\rmdefault}{\mddefault}{\updefault}{\color[rgb]{0,0,0}$\leftarrow aids(X)$}%
}}}
\put(2776,-511){\makebox(0,0)[lb]{\smash{\SetFigFont{5}{6.0}{\rmdefault}{\mddefault}{\updefault}{\color[rgb]{0,0,0}$X=p1$}%
}}}
\put(3301,-511){\makebox(0,0)[lb]{\smash{\SetFigFont{5}{6.0}{\rmdefault}{\mddefault}{\updefault}{\color[rgb]{0,0,0}$X=p3$}%
}}}
\put(3826,-511){\makebox(0,0)[lb]{\smash{\SetFigFont{5}{6.0}{\rmdefault}{\mddefault}{\updefault}{\color[rgb]{0,0,0}$X=p2$}%
}}}
\put(6526, 14){\makebox(0,0)[lb]{\smash{\SetFigFont{5}{6.0}{\rmdefault}{\mddefault}{\updefault}{\color[rgb]{0,0,0}${\cal C}_4$}%
}}}
\put(5851,-886){\makebox(0,0)[lb]{\smash{\SetFigFont{6}{7.2}{\rmdefault}{\mddefault}{\updefault}{\color[rgb]{0,0,0}$\leftarrow contact(X,p2)$}%
}}}
\put(7201,-886){\makebox(0,0)[lb]{\smash{\SetFigFont{6}{7.2}{\rmdefault}{\mddefault}{\updefault}{\color[rgb]{0,0,0}$\leftarrow contact(X,p3)$}%
}}}
\put(4426,-886){\makebox(0,0)[lb]{\smash{\SetFigFont{6}{7.2}{\rmdefault}{\mddefault}{\updefault}{\color[rgb]{0,0,0}$\leftarrow contact(X,p1)$}%
}}}
\put(7126,-661){\makebox(0,0)[lb]{\smash{\SetFigFont{5}{6.0}{\rmdefault}{\mddefault}{\updefault}{\color[rgb]{0,0,0}$Y=p3$}%
}}}
\end{picture}
\caption{The proof tree for $\leftarrow aids(X)$.} 
\label{slg-aids}
\end{figure}
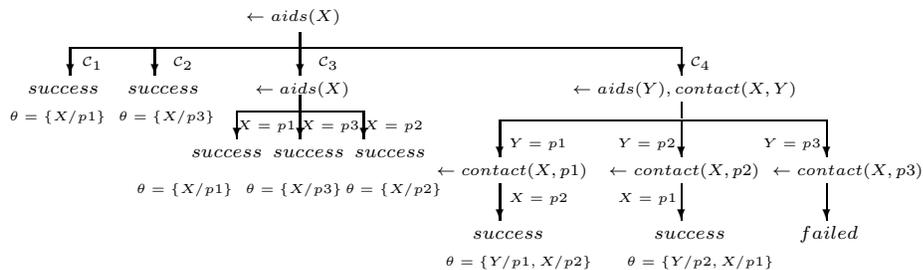
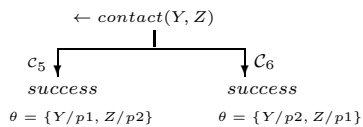
\begin{figure}[htb]
\begin{center}
\setlength{\unitlength}{3000sp}%
\begingroup\makeatletter\ifx\SetFigFont\undefined%
\gdef\SetFigFont#1#2#3#4#5{%
  \reset@font\fontsize{#1}{#2pt}%
  \fontfamily{#3}\fontseries{#4}\fontshape{#5}%
  \selectfont}%
\fi\endgroup%
\begin{picture}(2025,961)(1276,-392)
\thinlines
{\color[rgb]{0,0,0}\put(2476,389){\line( 0,-1){130}}
}%
{\color[rgb]{0,0,0}\put(1676,239){\vector( 0,-1){225}}
\put(1676,239){\line( 1, 0){1550}}
\put(3226,239){\vector( 0,-1){225}}
}%
\put(1801,464){\makebox(0,0)[lb]{\smash{\SetFigFont{6}{7.2}{\rmdefault}{\mddefault}{\updefault}{\color[rgb]{0,0,0}$\leftarrow contact(Y,Z)$}%
}}}
\put(3301, 89){\makebox(0,0)[lb]{\smash{\SetFigFont{7}{8.4}{\rmdefault}{\mddefault}{\updefault}{\color[rgb]{0,0,0}${\cal C}_6$}%
}}}
\put(1426,-136){\makebox(0,0)[lb]{\smash{\SetFigFont{7}{8.4}{\rmdefault}{\mddefault}{\updefault}{\color[rgb]{0,0,0}$success$}%
}}}
\put(3076,-136){\makebox(0,0)[lb]{\smash{\SetFigFont{7}{8.4}{\rmdefault}{\mddefault}{\updefault}{\color[rgb]{0,0,0}$success$}%
}}}
\put(1276,-361){\makebox(0,0)[lb]{\smash{\SetFigFont{5}{6.0}{\rmdefault}{\mddefault}{\updefault}{\color[rgb]{0,0,0}$\theta = \{Y/p1,Z/p2\}$}%
}}}
\put(3001,-361){\makebox(0,0)[lb]{\smash{\SetFigFont{5}{6.0}{\rmdefault}{\mddefault}{\updefault}{\color[rgb]{0,0,0}$\theta = \{Y/p2,Z/p1\}$}%
}}}
\put(1426, 89){\makebox(0,0)[lb]{\smash{\SetFigFont{5}{6.0}{\rmdefault}{\mddefault}{\updefault}{\color[rgb]{0,0,0}${\cal C}_5$}%
}}}
\end{picture}
\end{center}
\caption{The proof tree for $\leftarrow contact(Y,Z)$.} 
\label{slg-aids2}
\end{figure}

For the computational complexity, we observe that 
the cost of Algorithm 2 is dominated
by applying SLG-resolution to evaluate the goals in $GS_0$.
It has been shown that for a Datalog program $P$, 
the time complexity of computing the well-founded model $WF(P)$
is polynomial \cite{VRS91,vardi82}. More precisely,
the time complexity of
SLG-resolution is $O(|P|*N^{\Pi_P+1}*log N)$,
where $|P|$ is the number of clauses in $P$,
$\Pi_P$ is the maximum number of literals in the body of
a clause, and $N$,
the number of atoms of predicates in $P$ that are not variants 
of each other, is a polynomial in the number 
of ground unit clauses in $P$ \cite{chen96}. 

$PB\cup CB$ is a Datalog program except for the $member(X_i, DOM_i)$
predicates (see Definition \ref{kb}). 
Since each domain $DOM_i$ is a finite list of constants,
checking if $X_i$ is in $DOM_i$ takes time linear in the size of $DOM_i$. 
Let $K_1$ be the maximum number of $member(X_i, DOM_i)$
predicates used in a clause in $P$ and $K_2$ be the maximum
size of a domain $DOM_i$. Then the time of 
handling all $member(X_i, DOM_i)$ predicates
in a clause is bounded by $K_1*K_2$. Since each clause in $P$
is applied at most $N$ times in SLG-resolution,
the time of handling all $member(X_i, DOM_i)$s
in all clauses in $P$ is bounded by $|P|*N*K_1*K_2$. 
This is also a polynomial, hence SLG-resolution
computes the well-founded model $WF(PB\cup CB)$
in polynomial time. Therefore,
we have the following result. 

\begin{theorem}
\label{th-comp1}
The time complexity of Algorithm 2 is polynomial.
\end{theorem}

\subsection{Probability Distributions Induced by $KB$}
For any random variable $A$,
we use $pa(A)$ to denote the set of random variables
that have direct influences on $A$; namely
$pa(A)$ consists of random variables in the body
of all influence clauses whose head is $A$.
Assume that the probability distribution
${\bf P}(A|pa(A))$ is available (see Section \ref{sec_cpt}). 
Furthermore, we make the following {\em independence assumption}.

\begin{assumption}
\label{ind-ass}
{\em
For any random variable $A$,
we assume that given $pa(A)$, $A$ is probabilistically independent
of all random variables in ${\cal S}(KB)$ that are
not influenced by $A$. 
}
\end{assumption}

We define probability distributions induced by $KB$
in terms of whether there are cyclic influences.

\begin{definition}
\label{no-cycle}
{\em
When no cyclic influence occurs, the probability distribution induced by $KB$
is ${\bf P}({\cal S}(KB))$.  
}
\end{definition}

\begin{theorem}
\label{th-prob}
${\bf P}({\cal S}(KB)) = \prod_{A_i\in {\cal S}(KB)}{\bf P}(A_i|pa(A_i))$
under the independence assumption.  
\end{theorem}

\noindent {\bf Proof:} When no cyclic influence occurs, 
the random variables in ${\cal S}(KB)$ can be arranged in 
a partial order such that if $A_i$ is influenced by
$A_j$ then $j>i$. By the independence assumption, we have
${\bf P}({\cal S}(KB))$ $=$ ${\bf P}(\bigwedge_{A_i\in {\cal S}(KB)} A_i)$ $=$ 
${\bf P}(A_1 | \bigwedge_{i=2} A_i) * {\bf P}(\bigwedge_{i=2} A_i)$ $=$ 
${\bf P}(A_1 | pa(A_1)) * {\bf P}(A_2 | \bigwedge_{i=3} A_i) * {\bf P}(\bigwedge_{i=3} A_i)$ $=$ 
$...$ $=$
$\prod_{A_i\in {\cal S}(KB)}{\bf P}(A_i|pa(A_i))$ $\Box$ \\[.1in]
\indent
When there are cyclic influences, we cannot have a partial order
on ${\cal S}(KB)$. By Definition \ref{inf-by} and Theorem \ref{main-inf},
any cyclic influence, say ``$A_1$ is influenced by itself,"
must be resulted from a set of influence clauses in ${\cal I}_{clause}(KB)$
of the form
\begin{eqnarray}
\label{cyc-inf}
A_1 & \leftarrow & ..., A_2, ... \nonumber \\
A_2 & \leftarrow & ..., A_3, ... \nonumber \\
 & ...... &\\
A_n & \leftarrow & ..., A_1, ... \nonumber     
\end{eqnarray}
These influence clauses generate a chain (cycle) of direct influences
\begin{eqnarray}
\label{cyc-feedback}
A_1 \leftarrow A_2 \leftarrow A_3 \leftarrow  ... \leftarrow   A_n  \leftarrow A_1
\end{eqnarray}
which defines a feedback connection. Since a feedback system can be modeled by 
a two-slice DBN (see Section \ref{subsec-1-1}), 
the above influence clauses represent the same knowledge as the following ones do:
\begin{eqnarray}
\label{cyc-inf2}
A_1 & \leftarrow & ..., A_2, ... \nonumber \\
A_2 & \leftarrow & ..., A_3, ... \nonumber \\
 & ...... &\\
A_n & \leftarrow & ..., A_{1_{t-1}}, ... \nonumber     
\end{eqnarray}
Here the $A_i$s are state variables and $A_{1_{t-1}}$
is a state input variable. As a result, $A_1$ being influenced by itself
becomes $A_1$ being influenced by $A_{1_{t-1}}$.
By applying this transformation 
(from influence clauses (\ref{cyc-inf}) to (\ref{cyc-inf2})), we can get
rid of all cyclic influences and obtain
a {\em generalized set} ${\cal I}_{clause}(KB)_g$ of influence clauses 
from ${\cal I}_{clause}(KB)$.\footnote{Depending on
starting from which influence clause to generate an influence cycle, 
a different generalized set containing different state input variables
would be obtained. All of them
are equivalent in the sense that they define the same feedbacks 
(cycles of direct influences) and
can be unrolled into the same stationary DBN.}

\begin{example}[Example \ref{aids-eg-infcl} continued]
\label{aids-eg-infclg}
{\em
${\cal I}_{clause}(KB_1)$ can be transformed to the following
generalized set of influence clauses 
by introducing three state input variables $aids(p1)_{t-1}$, $aids(p2)_{t-1}$
and $aids(p3)_{t-1}$.
\begin{tabbing}
$\qquad {\cal I}_{clause}(KB_1)_g:$ \= 1. $aids(p1).$\\
\> 2. $aids(p3).$\\
\> 3. $aids(p1)\leftarrow aids(p1)_{t-1}.$\\
\> 3. $aids(p2)\leftarrow aids(p2)_{t-1}.$\\
\> 3. $aids(p3)\leftarrow aids(p3)_{t-1}.$\\
\> 4. $aids(p2)\leftarrow aids(p1)_{t-1}, contact(p2,p1).$\\
\> 4. $aids(p1)\leftarrow aids(p2), contact(p1,p2).$\\
\> 5. $contact(p1, p2).$\\
\> 6. $contact(p2, p1).$
\end{tabbing}
}
\end{example}

When there is no cyclic influence, $KB$ is a non-temporal model,
represented by ${\cal I}_{clause}(KB)$. When cyclic influences 
occur, however, $KB$ becomes a temporal model,
represented by ${\cal I}_{clause}(KB)_g$.  
Let ${\cal S}(KB)_g$ be ${\cal S}(KB)$ plus all state input variables 
introduced in ${\cal I}_{clause}(KB)_g$.

\begin{definition}
\label{with-cycle}
{\em
When there are cyclic influences, the probability distribution induced by $KB$
is ${\bf P}({\cal S}(KB)_g)$.  
}
\end{definition}

By extending the independence assumption from ${\cal S}(KB)$ to ${\cal S}(KB)_g$,
we obtain the following result.

\begin{theorem}
\label{th-prob2}
${\bf P}({\cal S}(KB)_g) = \prod_{A_i\in {\cal S}(KB)_g}{\bf P}(A_i|pa(A_i))$
under the independence assumption.  
\end{theorem}

\noindent {\bf Proof:} Since ${\cal I}_{clause}(KB)_g$
produces no cyclic influences, 
the random variables in ${\cal S}(KB)_g$ can be arranged in 
a partial order such that if $A_i$ is influenced by
$A_j$ then $j>i$. The proof then proceeds in the same way as that of 
Theorem \ref{th-prob}. $\Box$

\section{Building a Bayesian Network from a Bayesian Knowledge Base}

\subsection{Building a Two-Slice DBN Structure}

From a Bayesian knowledge base $KB$, we can derive a set
of influence clauses ${\cal I}_{clause}(KB)$, which defines the same direct influence
relation over the same space ${\cal S}(KB)$ of random 
variables as $PB\cup CB$ does (see Theorem \ref{main-inf}). 
Therefore, given a probabilistic query 
together with some evidences, we can depict a network structure 
from ${\cal I}_{clause}(KB)$, which covers 
the random variables in the query and evidences,
by backward-chaining the related random variables 
via the direct influence relation. 

Let $Q$ be a probabilistic query and $E$ a set of evidences, where 
all random variables come from ${\cal S}(KB)$ (i.e., they 
are heads of some influence clauses in ${\cal I}_{clause}(KB)$).  
Let $TOP$ consist of these random variables.
An {\em influence network} of $Q$ and $E$,\footnote{Note the
differences between influence networks and {\em influence diagrams}.
Influence diagrams (also known as decision networks) are a formalism
introduced in decision theory that extends Bayesian networks by incorporating
actions and utilities \cite{RN95}.}
denoted ${\cal I}_{net}(KB)_{Q,E}$,
is constructed from ${\cal I}_{clause}(KB)$ using the following algorithm.\\[.1in]
{\bf Algorithm 3: Building an influence network.} 
\begin{enumerate}
\item
\label{alg3-item-1}
Initially, ${\cal I}_{net}(KB)_{Q,E}$ has all random variables in
$TOP$ as nodes.

\item
\label{alg3-item-2}
Remove the first random variable $A$ from $TOP$.
For each influence clause in ${\cal I}_{clause}(KB)$ of the form
$k.\ A\leftarrow A_1, ..., A_l$, if $l=0$ then 
add to ${\cal I}_{net}(KB)_{Q,E}$ an edge $A \stackrel{k}{\leftarrow}$. Otherwise,
for each $A_i$ in the body 
\begin{enumerate}
\item
\label{alg3-item-2a}
If $A_i$ is not in ${\cal I}_{net}(KB)_{Q,E}$ then add $A_i$ 
to ${\cal I}_{net}(KB)_{Q,E}$ as a new node and add it to the end of $TOP$.
\item
\label{alg3-item-2b}
Add to ${\cal I}_{net}(KB)_{Q,E}$ an edge $A \stackrel{k}{\leftarrow} A_i$.
\end{enumerate}

\item 
Repeat step 2 until $TOP$ becomes empty.

\item
Return ${\cal I}_{net}(KB)_{Q,E}$.
\end{enumerate}

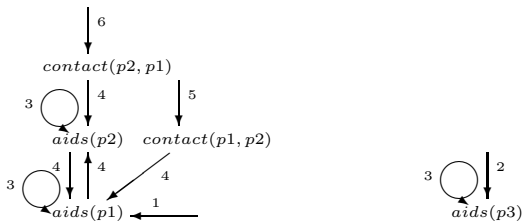
\begin{figure*}[htb]
\centering
\setlength{\unitlength}{3000sp}%
\begingroup\makeatletter\ifx\SetFigFont\undefined%
\gdef\SetFigFont#1#2#3#4#5{%
  \reset@font\fontsize{#1}{#2pt}%
  \fontfamily{#3}\fontseries{#4}\fontshape{#5}%
  \selectfont}%
\fi\endgroup%
\begin{picture}(4050,1779)(2926,-2878)
\thinlines
{\color[rgb]{0,0,0}\put(4501,-2836){\vector(-1, 0){550}}
}%
\put(3226,-1636){\makebox(0,0)[lb]{\smash{\SetFigFont{6}{7.2}{\rmdefault}{\mddefault}{\updefault}{\color[rgb]{0,0,0}$contact(p2,p1)$}%
}}}
\put(3301,-2236){\makebox(0,0)[lb]{\smash{\SetFigFont{6}{7.2}{\rmdefault}{\mddefault}{\updefault}{\color[rgb]{0,0,0}$aids(p2)$}%
}}}
\put(4051,-2236){\makebox(0,0)[lb]{\smash{\SetFigFont{6}{7.2}{\rmdefault}{\mddefault}{\updefault}{\color[rgb]{0,0,0}$contact(p1,p2)$}%
}}}
\put(3676,-1861){\makebox(0,0)[lb]{\smash{\SetFigFont{5}{6.0}{\rmdefault}{\mddefault}{\updefault}{\color[rgb]{0,0,0}$4$}%
}}}
\put(3676,-1261){\makebox(0,0)[lb]{\smash{\SetFigFont{5}{6.0}{\rmdefault}{\mddefault}{\updefault}{\color[rgb]{0,0,0}$6$}%
}}}
\put(4426,-1861){\makebox(0,0)[lb]{\smash{\SetFigFont{5}{6.0}{\rmdefault}{\mddefault}{\updefault}{\color[rgb]{0,0,0}$5$}%
}}}
\put(3301,-2836){\makebox(0,0)[lb]{\smash{\SetFigFont{6}{7.2}{\rmdefault}{\mddefault}{\updefault}{\color[rgb]{0,0,0}$aids(p1)$}%
}}}
\put(3301,-2461){\makebox(0,0)[lb]{\smash{\SetFigFont{5}{6.0}{\rmdefault}{\mddefault}{\updefault}{\color[rgb]{0,0,0}$4$}%
}}}
\put(3676,-2461){\makebox(0,0)[lb]{\smash{\SetFigFont{5}{6.0}{\rmdefault}{\mddefault}{\updefault}{\color[rgb]{0,0,0}$4$}%
}}}
\put(4201,-2536){\makebox(0,0)[lb]{\smash{\SetFigFont{5}{6.0}{\rmdefault}{\mddefault}{\updefault}{\color[rgb]{0,0,0}$4$}%
}}}
\put(4126,-2761){\makebox(0,0)[lb]{\smash{\SetFigFont{5}{6.0}{\rmdefault}{\mddefault}{\updefault}{\color[rgb]{0,0,0}$1$}%
}}}
\put(3076,-1936){\makebox(0,0)[lb]{\smash{\SetFigFont{5}{6.0}{\rmdefault}{\mddefault}{\updefault}{\color[rgb]{0,0,0}$3$}%
}}}
\put(2926,-2611){\makebox(0,0)[lb]{\smash{\SetFigFont{5}{6.0}{\rmdefault}{\mddefault}{\updefault}{\color[rgb]{0,0,0}$3$}%
}}}
{\color[rgb]{0,0,0}\put(3451,-2311){\vector( 0,-1){375}}
}%
{\color[rgb]{0,0,0}\put(3368,-1936){\circle{316}}
}%
{\color[rgb]{0,0,0}\put(3376,-2096){\vector( 3,-2){ 72.692}}
}%
{\color[rgb]{0,0,0}\put(3218,-2611){\circle{316}}
}%
{\color[rgb]{0,0,0}\put(3226,-2771){\vector( 3,-2){ 72.692}}
}%
{\color[rgb]{0,0,0}\put(6668,-2536){\circle{316}}
}%
{\color[rgb]{0,0,0}\put(6676,-2696){\vector( 3,-2){ 72.692}}
}%
{\color[rgb]{0,0,0}\put(6901,-2311){\vector( 0,-1){375}}
}%
\put(6601,-2836){\makebox(0,0)[lb]{\smash{\SetFigFont{6}{7.2}{\rmdefault}{\mddefault}{\updefault}{\color[rgb]{0,0,0}$aids(p3)$}%
}}}
\put(6976,-2461){\makebox(0,0)[lb]{\smash{\SetFigFont{5}{6.0}{\rmdefault}{\mddefault}{\updefault}{\color[rgb]{0,0,0}$2$}%
}}}
\put(6376,-2536){\makebox(0,0)[lb]{\smash{\SetFigFont{5}{6.0}{\rmdefault}{\mddefault}{\updefault}{\color[rgb]{0,0,0}$3$}%
}}}
{\color[rgb]{0,0,0}\put(3601,-1711){\vector( 0,-1){375}}
}%
{\color[rgb]{0,0,0}\put(3601,-1111){\vector( 0,-1){375}}
}%
{\color[rgb]{0,0,0}\put(4351,-1711){\vector( 0,-1){375}}
}%
{\color[rgb]{0,0,0}\put(4276,-2311){\vector(-4,-3){516}}
}%
{\color[rgb]{0,0,0}\put(3601,-2686){\vector( 0, 1){375}}
}%
\end{picture}
\caption{An influence network built from the AIDS program $KB_1$.} 
\label{aids-inf-net}
\end{figure*}

\begin{example}[Example \ref{aids-eg-infcl} continued]
\label{aids-eg-inf-net}
{\em
To build an influence network from $KB_1$
that covers $aids(p1)$, $aids(p2)$ and $aids(p3)$, we apply Algorithm 3 
to ${\cal I}_{clause}($ $KB_1)$ while letting
$TOP = \{aids(p1), aids(p2), aids(p3)\}$. It generates an influence network
${\cal I}_{net}(KB_1)_{Q,E}$ as shown in Figure \ref{aids-inf-net}.
}
\end{example}

An influence network is a graphical representation for influence clauses. 
This claim is supported by the following properties
of influence networks.

\begin{theorem}
\label{th-alg3}
For any $A_i, A_j$ in ${\cal I}_{net}(KB)_{Q,E}$,
$A_j$ is a parent node of $A_i$, connected via
an edge $A_i \stackrel{k}{\leftarrow} A_j$, 
if and only if there is an influence clause of the form 
$k.\ \ A_i\leftarrow A_1,...,A_j,...,A_l$ in ${\cal I}_{clause}(KB)$.
\end{theorem}

\noindent {\bf Proof:}
First note that termination of Algorithm 3 is guaranteed by the
fact that any random variable in ${\cal S}(KB)$ will be added to $TOP$
no more than one time (line \ref{alg3-item-2a}). 
Let $A_i, A_j$ be nodes in ${\cal I}_{net}(KB)_{Q,E}$. If 
$A_j$ is a parent node of $A_i$, connected via
an edge $A_i \stackrel{k}{\leftarrow} A_j$, this edge must be
added at line \ref{alg3-item-2b}, due to applying  
an influence clause in ${\cal I}_{clause}(KB)$ of the form 
$k.\ \ A_i\leftarrow A_1,...,A_j,...,A_l$ (line \ref{alg3-item-2}).
Conversely, if ${\cal I}_{clause}(KB)$ contains such an influence clause,
it must be applied at line \ref{alg3-item-2}, with edges of the form
$A_i \stackrel{k}{\leftarrow} A_j$ added to the network
at line \ref{alg3-item-2b}. $\Box$

\begin{theorem}
\label{th-alg3-2}
For any $A_i, A_j$ in ${\cal I}_{net}(KB)_{Q,E}$,
$A_i$ is a descendant node of $A_j$ if and only if $A_i$ is influenced by $A_j$.
\end{theorem}

\noindent {\bf Proof:} Assume $A_i$ is a descendant node of $A_j$, with a path
\begin{equation}
\label{eq-k}
A_i\stackrel{k}{\leftarrow} B_1\stackrel{k_1}{\leftarrow} ... B_m \stackrel{k_m}{\leftarrow} A_j
\end{equation}
By Theorem \ref{th-alg3}, ${\cal I}_{clause}(KB)$ must contain the following influence clauses
\begin{eqnarray}
\label{alg3-inf-chain}
k. & A_i\leftarrow ..., B_1,... \nonumber \\
k_1. & B_1\leftarrow ..., B_2,... \nonumber \\
 & ...... \\
k_m. & B_m\leftarrow ..., A_j,... \nonumber     
\end{eqnarray}
By Theorem \ref{main-inf} and Definition \ref{inf-by},
$A_i$ is influenced by $A_j$. Conversely, 
if $A_i$ is influenced by $A_j$,
there must be a chain of influence clauses of the form as above.
Since $A_i, A_j$ are in ${\cal I}_{net}(KB)_{Q,E}$, by Theorem \ref{th-alg3}
there must be a path of form (\ref{eq-k}) in the network. $\Box$

\begin{theorem}
\label{th-alg3-3}
Let $V$ be the set of nodes in ${\cal I}_{net}(KB)_{Q,E}$ 
and let $W = \{A_j \in {\cal S}(KB) |$ for some $A_i\in TOP$,
$A_i$ is influenced by $A_j\}$. $V = TOP \cup W$.\footnote{This result
suggests that an influence network is similar to a {\em supporting network}
introduced in \cite{ngo-had97}.}
\end{theorem} 

\noindent {\bf Proof:}
That ${\cal I}_{net}(KB)_{Q,E}$ covers all random variables in $TOP$
follows from line \ref{alg3-item-1} of Algorithm 3.
We first prove that if $A_j\in W$ then $A_j\in V$.
Assume $A_j\in W$. There must be a chain of influence clauses of form (\ref{alg3-inf-chain})
with $A_i\in TOP$. In this case, $B_1, B_2, ..., B_m, A_j$ will be recursively added 
to the network (line \ref{alg3-item-2}). Thus $A_j\in V$.
We then prove that if $A_j\in V$ and $A_j\not\in TOP$ then $A_j\in W$.
Assume $A_j\in V$ and $A_j\not\in TOP$. $A_j$ must not be added to $V$
at line \ref{alg3-item-1}. Instead, it is added to $V$ at line 
\ref{alg3-item-2a}. This means that for some $A_i\in TOP$, $A_i$ is a descendant
of $A_j$. By Theorem \ref{th-alg3-2}, $A_i$ is influenced by $A_j$.
Hence $A_j\in W$. $\Box$\\[.1in]
\indent
Theorem \ref{th-prob} shows that the probability distribution induced by $KB$
can be computed over ${\cal I}_{clause}(KB)$.
Let ${\cal I}_{net}(KB)_{{\cal S}(KB)}$ denote an influence network
that covers all random variables in ${\cal S}(KB)$.
We show that the same distribution can be computed over  
${\cal I}_{net}(KB)_{{\cal S}(KB)}$. 
For any node $A_i$ in ${\cal I}_{net}(KB)_{{\cal S}(KB)}$, let
$parents(A_i)$ denote the set of parent nodes of $A_i$ in the network.
Observe the following facts: First, by Theorem \ref{th-alg3}, $parents(A_i) = pa(A_i)$.
Second, by Theorem \ref{th-alg3-2},
$A_i$ is a descendant node of $A_j$ in ${\cal I}_{net}(KB)_{{\cal S}(KB)}$ 
if and only if $A_i$ is influenced by $A_j$ in ${\cal I}_{clause}(KB)$.
This means that the independence assumption 
(Assumption \ref{ind-ass}) applies to ${\cal I}_{net}(KB)_{{\cal S}(KB)}$ as well,
and that ${\cal I}_{clause}(KB)$ produces a cycle of direct influences if and only if
${\cal I}_{net}(KB)_{{\cal S}(KB)}$ contains the same (direct) loop. 
Combining these facts leads to the following immediate result. 

\begin{theorem}
\label{th-prob-1}
When no cyclic influence occurs, the probability distribution induced by $KB$
can be computed over ${\cal I}_{net}(KB)_{{\cal S}(KB)}$.
That is, ${\bf P}({\cal S}(KB))$ $=$ $\prod_{A_i\in {\cal S}(KB)}{\bf P}(A_i|pa(A_i))$
$=$ $\prod_{A_i\in {\cal S}(KB)}{\bf P}(A_i|parents(A_i))$
under the independence assumption.    
\end{theorem}

Theorem \ref{th-prob-1} implies that an influence network without loops 
is a Bayesian network structure.
Let us consider influence networks with loops. By Theorem \ref{th-alg3-2},
loops in an influence network are generated from recursive influence
clauses of form (\ref{cyc-inf}) and thus they depict feedback connections 
of form (\ref{cyc-feedback}). This means that an influence network with loops
can be converted into a two-slice DBN, simply by converting each loop
of the form
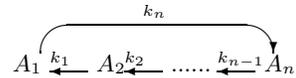
\begin{center}
\setlength{\unitlength}{3079sp}%
\begingroup\makeatletter\ifx\SetFigFont\undefined%
\gdef\SetFigFont#1#2#3#4#5{%
  \reset@font\fontsize{#1}{#2pt}%
  \fontfamily{#3}\fontseries{#4}\fontshape{#5}%
  \selectfont}%
\fi\endgroup%
\begin{picture}(2108,597)(751,-553)
\thinlines
{\color[rgb]{0,0,0}\put(2851,-361){\vector( 0,-1){0}}
\put(1914,-361){\oval(1874,450)[tr]}
\put(1914,-361){\oval(1876,450)[tl]}
}%
{\color[rgb]{0,0,0}\put(1951,-511){\vector(-1, 0){300}}
}%
{\color[rgb]{0,0,0}\put(2701,-511){\vector(-1, 0){300}}
}%
{\color[rgb]{0,0,0}\put(1351,-511){\vector(-1, 0){300}}
}%
\put(2026,-511){\makebox(0,0)[lb]{\smash{\SetFigFont{9}{10.8}{\rmdefault}{\mddefault}{\updefault}{\color[rgb]{0,0,0}......}%
}}}
\put(2776,-511){\makebox(0,0)[lb]{\smash{\SetFigFont{9}{10.8}{\rmdefault}{\mddefault}{\updefault}{\color[rgb]{0,0,0}$A_n$}%
}}}
\put(751,-511){\makebox(0,0)[lb]{\smash{\SetFigFont{9}{10.8}{\rmdefault}{\mddefault}{\updefault}{\color[rgb]{0,0,0}$A_1$}%
}}}
\put(1426,-511){\makebox(0,0)[lb]{\smash{\SetFigFont{9}{10.8}{\rmdefault}{\mddefault}{\updefault}{\color[rgb]{0,0,0}$A_2$}%
}}}
\put(1051,-436){\makebox(0,0)[lb]{\smash{\SetFigFont{6}{7.2}{\rmdefault}{\mddefault}{\updefault}{\color[rgb]{0,0,0}$k_1$}%
}}}
\put(1651,-436){\makebox(0,0)[lb]{\smash{\SetFigFont{6}{7.2}{\rmdefault}{\mddefault}{\updefault}{\color[rgb]{0,0,0}$k_2$}%
}}}
\put(2401,-436){\makebox(0,0)[lb]{\smash{\SetFigFont{6}{7.2}{\rmdefault}{\mddefault}{\updefault}{\color[rgb]{0,0,0}$k_{n-1}$}%
}}}
\put(1801,-61){\makebox(0,0)[lb]{\smash{\SetFigFont{6}{7.2}{\rmdefault}{\mddefault}{\updefault}{\color[rgb]{0,0,0}$k_n$}%
}}}
\end{picture}
\end{center}
into a two-slice DBN path
\[A_1 \stackrel{k_1}{\leftarrow} A_2 \stackrel{k_2}{\leftarrow} 
... \stackrel{k_{n-1}}{\leftarrow} A_n\stackrel{k_n}{\leftarrow} 
A_{1_{t-1}}\]  
by introducing a state input node $A_{1_{t-1}}$.

As illustrated in Section \ref{subsec-1-1}, a two-slice DBN is a snapshot of a stationary
DBN across any two time slices, which can be obtained by traversing the stationary DBN from 
a set of state variables backward to the same set of state variables (i.e., state input nodes).
This process corresponds to generating an influence network ${\cal I}_{net}(KB)_{Q,E}$ from
${\cal I}_{clause}(KB)$ incrementally (adding nodes and edges one at a time) while 
wrapping up loop nodes with state input nodes. This leads to the following algorithm
for building a two-slice DBN structure,
$2{\cal S}_{net}(KB)_{Q,E}$, directly from ${\cal I}_{clause}(KB)$,
where $Q$, $E$ and $TOP$ are the same as defined in Algorithm 3.\\[.1in]
{\bf Algorithm 4: Building a two-slice DBN structure.} 
\begin{enumerate}
\item
Initially, $2{\cal S}_{net}(KB)_{Q,E}$ has all random variables in
$TOP$ as nodes.

\item
Remove the first random variable $A$ from $TOP$.
For each influence clause in ${\cal I}_{clause}(KB)$ of the form
$k.\ \ A\leftarrow A_1, ..., A_l$, if $l=0$ then 
add to $2{\cal S}_{net}(KB)_{Q,E}$ an edge $A \stackrel{k}{\leftarrow}$. Otherwise,
for each $A_i$ in the body 
\begin{enumerate}
\item
If $A_i$ is not in $2{\cal S}_{net}(KB)_{Q,E}$ then add $A_i$
to $2{\cal S}_{net}(KB)_{Q,E}$ as a new node and add it to the end of $TOP$.
\item
\label{loopcut}
If adding $A \stackrel{k}{\leftarrow} A_i$ to $2{\cal S}_{net}(KB)_{Q,E}$ produces a loop, 
then add to $2{\cal S}_{net}(KB)_{Q,E}$ a node $A_{i_{t-1}}$
and an edge $A \stackrel{k}{\leftarrow} A_{i_{t-1}}$,
else add an edge $A \stackrel{k}{\leftarrow} A_i$ to $2{\cal S}_{net}(KB)_{Q,E}$.
\end{enumerate}

\item 
Repeat step 2 until $TOP$ becomes empty.

\item
Return $2{\cal S}_{net}(KB)_{Q,E}$.
\end{enumerate} 

\begin{example}[Example \ref{aids-eg-inf-net} continued]
\label{aids-eg2}
{\em
To build a two-slice DBN structure from $KB_1$
that covers $aids(p1)$, $aids(p2)$ and $aids(p3)$, we apply Algorithm 4 
to ${\cal I}_{clause}($ $KB_1)$ while letting
$TOP = \{aids(p1), aids(p2), aids(p3)\}$. It generates 
$2{\cal S}_{net}(KB_1)_{Q,E}$ as shown in Figure \ref{aids}.
Note that loops are cut
by introducing three state input nodes $aids(p1)_{t-1}$,
$aids(p2)_{t-1}$ and $aids(p3)_{t-1}$.
The two-slice DBN structure 
concisely depicts a feedback system where the feedback
connections are as shown in Figure \ref{feedback1}.
\begin{figure*}[htb]
\centering
\setlength{\unitlength}{3000sp}%
\begingroup\makeatletter\ifx\SetFigFont\undefined%
\gdef\SetFigFont#1#2#3#4#5{%
  \reset@font\fontsize{#1}{#2pt}%
  \fontfamily{#3}\fontseries{#4}\fontshape{#5}%
  \selectfont}%
\fi\endgroup%
\begin{picture}(4875,1779)(2101,-2878)
\put(3676,-1261){\makebox(0,0)[lb]{\smash{\SetFigFont{5}{6.0}{\rmdefault}{\mddefault}{\updefault}{\color[rgb]{0,0,0}$6$}%
}}}
\put(4426,-1861){\makebox(0,0)[lb]{\smash{\SetFigFont{5}{6.0}{\rmdefault}{\mddefault}{\updefault}{\color[rgb]{0,0,0}$5$}%
}}}
\put(3676,-2461){\makebox(0,0)[lb]{\smash{\SetFigFont{5}{6.0}{\rmdefault}{\mddefault}{\updefault}{\color[rgb]{0,0,0}$4$}%
}}}
\put(4201,-2536){\makebox(0,0)[lb]{\smash{\SetFigFont{5}{6.0}{\rmdefault}{\mddefault}{\updefault}{\color[rgb]{0,0,0}$4$}%
}}}
\put(4126,-2761){\makebox(0,0)[lb]{\smash{\SetFigFont{5}{6.0}{\rmdefault}{\mddefault}{\updefault}{\color[rgb]{0,0,0}$1$}%
}}}
\put(2776,-2536){\makebox(0,0)[lb]{\smash{\SetFigFont{5}{6.0}{\rmdefault}{\mddefault}{\updefault}{\color[rgb]{0,0,0}$3$}%
}}}
\put(2101,-1636){\makebox(0,0)[lb]{\smash{\SetFigFont{6}{7.2}{\rmdefault}{\mddefault}{\updefault}{\color[rgb]{0,0,0}$aids(p2)_{t-1}$}%
}}}
\put(2776,-1936){\makebox(0,0)[lb]{\smash{\SetFigFont{5}{6.0}{\rmdefault}{\mddefault}{\updefault}{\color[rgb]{0,0,0}$3$}%
}}}
\put(5626,-2236){\makebox(0,0)[lb]{\smash{\SetFigFont{6}{7.2}{\rmdefault}{\mddefault}{\updefault}{\color[rgb]{0,0,0}$aids(p3)_{t-1}$}%
}}}
\put(3001,-2161){\makebox(0,0)[lb]{\smash{\SetFigFont{5}{6.0}{\rmdefault}{\mddefault}{\updefault}{\color[rgb]{0,0,0}$4$}%
}}}
\put(6526,-2836){\makebox(0,0)[lb]{\smash{\SetFigFont{6}{7.2}{\rmdefault}{\mddefault}{\updefault}{\color[rgb]{0,0,0}$aids(p3)$}%
}}}
\put(3301,-2836){\makebox(0,0)[lb]{\smash{\SetFigFont{6}{7.2}{\rmdefault}{\mddefault}{\updefault}{\color[rgb]{0,0,0}$aids(p1)$}%
}}}
\put(2101,-2236){\makebox(0,0)[lb]{\smash{\SetFigFont{6}{7.2}{\rmdefault}{\mddefault}{\updefault}{\color[rgb]{0,0,0}$aids(p1)_{t-1}$}%
}}}
\put(6076,-2536){\makebox(0,0)[lb]{\smash{\SetFigFont{5}{6.0}{\rmdefault}{\mddefault}{\updefault}{\color[rgb]{0,0,0}$3$}%
}}}
\put(6976,-2536){\makebox(0,0)[lb]{\smash{\SetFigFont{5}{6.0}{\rmdefault}{\mddefault}{\updefault}{\color[rgb]{0,0,0}$2$}%
}}}
\thinlines
{\color[rgb]{0,0,0}\put(3601,-1711){\vector( 0,-1){375}}
}%
{\color[rgb]{0,0,0}\put(3601,-1111){\vector( 0,-1){375}}
}%
{\color[rgb]{0,0,0}\put(4351,-1711){\vector( 0,-1){375}}
}%
{\color[rgb]{0,0,0}\put(4276,-2311){\vector(-4,-3){516}}
}%
{\color[rgb]{0,0,0}\put(3601,-2311){\vector( 0,-1){375}}
}%
{\color[rgb]{0,0,0}\put(4501,-2836){\vector(-1, 0){550}}
}%
{\color[rgb]{0,0,0}\put(2694,-1724){\vector( 2,-1){690}}
}%
{\color[rgb]{0,0,0}\put(2960,-2236){\vector( 1, 0){326}}
}%
{\color[rgb]{0,0,0}\put(2701,-2311){\vector( 2,-1){690}}
}%
{\color[rgb]{0,0,0}\put(6901,-2311){\vector( 0,-1){375}}
}%
{\color[rgb]{0,0,0}\put(6001,-2311){\vector( 2,-1){690}}
}%
\put(3226,-1636){\makebox(0,0)[lb]{\smash{\SetFigFont{6}{7.2}{\rmdefault}{\mddefault}{\updefault}{\color[rgb]{0,0,0}$contact(p2,p1)$}%
}}}
\put(3301,-2236){\makebox(0,0)[lb]{\smash{\SetFigFont{6}{7.2}{\rmdefault}{\mddefault}{\updefault}{\color[rgb]{0,0,0}$aids(p2)$}%
}}}
\put(4051,-2236){\makebox(0,0)[lb]{\smash{\SetFigFont{6}{7.2}{\rmdefault}{\mddefault}{\updefault}{\color[rgb]{0,0,0}$contact(p1,p2)$}%
}}}
\put(3676,-1861){\makebox(0,0)[lb]{\smash{\SetFigFont{5}{6.0}{\rmdefault}{\mddefault}{\updefault}{\color[rgb]{0,0,0}$4$}%
}}}
\end{picture}
\caption{A two-slice DBN structure built from the AIDS program $KB_1$.} 
\label{aids}
\end{figure*}
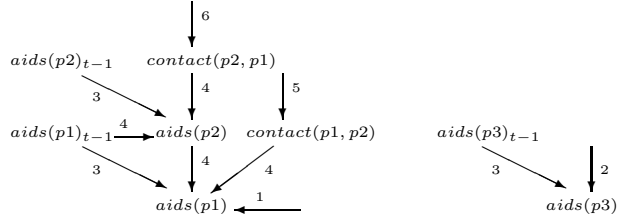
\begin{figure*}[htb]
\centering
\setlength{\unitlength}{3000sp}%
\begingroup\makeatletter\ifx\SetFigFont\undefined%
\gdef\SetFigFont#1#2#3#4#5{%
  \reset@font\fontsize{#1}{#2pt}%
  \fontfamily{#3}\fontseries{#4}\fontshape{#5}%
  \selectfont}%
\fi\endgroup%
\begin{picture}(6699,924)(589,-1873)
\put(6291,-1461){\makebox(0,0)[lb]{\smash{\SetFigFont{6}{7.2}{\familydefault}{\mddefault}{\updefault}{\color[rgb]{0,0,0}$aids(p3)$}%
}}}
\put(5101,-1336){\makebox(0,0)[lb]{\smash{\SetFigFont{6}{7.2}{\familydefault}{\mddefault}{\updefault}{\color[rgb]{0,0,0}$aids(p3)_{t-1}$}%
}}}
\put(901,-1486){\makebox(0,0)[lb]{\smash{\SetFigFont{6}{7.2}{\familydefault}{\mddefault}{\updefault}{\color[rgb]{0,0,0}$aids(p1)_{t-1}$}%
}}}
\put(3001,-1306){\makebox(0,0)[lb]{\smash{\SetFigFont{6}{7.2}{\familydefault}{\mddefault}{\updefault}{\color[rgb]{0,0,0}$aids(p1)$}%
}}}
\put(2021,-1306){\makebox(0,0)[lb]{\smash{\SetFigFont{6}{7.2}{\familydefault}{\mddefault}{\updefault}{\color[rgb]{0,0,0}$aids(p2)$}%
}}}
\put(901,-1186){\makebox(0,0)[lb]{\smash{\SetFigFont{6}{7.2}{\familydefault}{\mddefault}{\updefault}{\color[rgb]{0,0,0}$aids(p2)_{t-1}$}%
}}}
\thicklines
{\color[rgb]{0,0,0}\put(5986,-1671){\framebox(1115,510){}}
}%
\thinlines
{\color[rgb]{0,0,0}\multiput(7276,-1411)(0.00000,-180.00000){3}{\line( 0,-1){ 90.000}}
\multiput(7276,-1861)(-155.17241,0.00000){15}{\line(-1, 0){ 77.586}}
\multiput(5026,-1861)(0.00000,154.00000){3}{\line( 0, 1){ 77.000}}
\put(5026,-1476){\vector( 0, 1){0}}
}%
{\color[rgb]{0,0,0}\multiput(6901,-1411)(150.00000,0.00000){3}{\line( 1, 0){ 75.000}}
}%
{\color[rgb]{0,0,0}\put(4801,-1411){\vector( 1, 0){1400}}
}%
{\color[rgb]{0,0,0}\put(601,-1586){\line( 1, 0){2571}}
\put(3172,-1586){\vector( 0, 1){205}}
}%
{\color[rgb]{0,0,0}\multiput(3976,-1261)(0.00000,-171.42857){4}{\line( 0,-1){ 85.714}}
\multiput(3976,-1861)(-161.53846,0.00000){20}{\line(-1, 0){ 80.769}}
\multiput(826,-1861)(0.00000,190.00000){2}{\line( 0, 1){ 95.000}}
\put(826,-1576){\vector( 0, 1){0}}
}%
\thicklines
{\color[rgb]{0,0,0}\put(1786,-1771){\framebox(2015,710){}}
}%
\thinlines
{\color[rgb]{0,0,0}\put(2326,-1586){\vector( 0, 1){215}}
}%
{\color[rgb]{0,0,0}\put(2621,-1261){\vector( 1, 0){370}}
}%
{\color[rgb]{0,0,0}\multiput(2751,-1261)(0.00000,200.00000){2}{\line( 0, 1){100.000}}
\multiput(2751,-961)(-154.00000,0.00000){13}{\line(-1, 0){ 77.000}}
\multiput(826,-961)(0.00000,-200.00000){2}{\line( 0,-1){100.000}}
\put(826,-1261){\vector( 0,-1){0}}
}%
{\color[rgb]{0,0,0}\multiput(3601,-1261)(150.00000,0.00000){3}{\line( 1, 0){ 75.000}}
}%
{\color[rgb]{0,0,0}\put(601,-1261){\vector( 1, 0){1400}}
}%
\end{picture}
\caption{The feedback connections created by the AIDS program $KB_1$.} 
\label{feedback1}
\end{figure*}
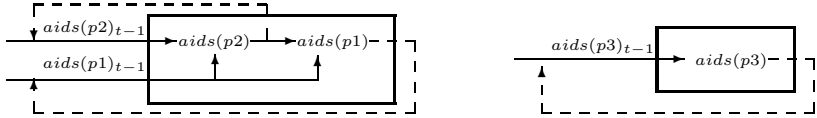
}
\end{example}

Algorithm 4 is Algorithm 3 enhanced with a mechanism for cutting loops 
(item \ref{loopcut}), i.e. when adding 
the current edge $A \stackrel{k}{\leftarrow} A_i$ to the network 
forms a loop, we replace it with an edge $A \stackrel{k}{\leftarrow} A_{i_{t-1}}$, 
where $A_{i_{t-1}}$ is a state input node. This is a process of
transforming influence clauses (\ref{cyc-inf}) to (\ref{cyc-inf2}). Therefore,
$2{\cal S}_{net}(KB)_{Q,E}$ can be viewed as an influence network built from
a generalized set ${\cal I}_{clause}(KB)_g$ of influence clauses.
Let ${\cal S}(KB)_g$ be the set of random variables in 
${\cal I}_{clause}(KB)_g$, as defined in Theorem \ref{th-prob2}.
Let $2{\cal S}_{net}(KB)_{{\cal S}(KB)}$ denote a two-slice DBN 
structure (produced by applying Algorithm 4) 
that covers all random variables in ${\cal S}(KB)_g$.
We then have the following immediate result from 
Theorem \ref{th-prob-1}.

\begin{theorem}
\label{th-prob-2}
When ${\cal I}_{clause}(KB)$ produces cyclic influences, 
the probability distribution induced by $KB$
can be computed over $2{\cal S}_{net}(KB)_{{\cal S}(KB)}$.
That is, ${\bf P}({\cal S}(KB)_g)$ $=$ $\prod_{A_i\in {\cal S}(KB)_g}{\bf P}(A_i$ $|pa(A_i))$
$=$ $\prod_{A_i\in {\cal S}(KB)_g}{\bf P}(A_i|parents(A_i))$
under the independence assumption.    
\end{theorem}

\begin{remark}
{\em
Note that Algorithm 4 produces a DBN structure without using any explicit time
parameters. It only requires the user to specify, via the query and evidences,
what random variables are necessarily included in the network.
Algorithm 4 builds a two-slice DBN structure for any given 
query and evidences whose random variables are heads of some 
influence clauses in ${\cal I}_{clause}(KB)$. 
When no query and evidences are provided, we may apply 
Algorithm 4 to build a {\em complete} two-slice DBN structure, 
$2{\cal S}_{net}(KB)_{{\cal S}(KB)}$, which covers 
the space ${\cal S}(KB)$ of random variables,   
by letting $TOP$ consist of all heads of influence clauses 
in ${\cal I}_{clause}(KB)$. This is a very useful feature, as
in many situations the user may not be able to 
present the right queries unless a Bayesian network structure is shown.

Also note that when there is no cyclic influence, 
Algorithm 4 becomes Algorithm 3 and thus it builds a 
regular Bayesian network structure. 
}
\end{remark}

\subsection{Building CPTs}
\label{sec_cpt}
After a Bayesian network structure $2{\cal S}_{net}(KB)_{Q,E}$  
has been constructed from a Bayesian knowledge base $KB$, 
we associate each (non-state-input) node $A$ in the network with a CPT. 
There are three cases. (1) If $A$ (as a head) only has unit clauses in ${\cal I}_{clause}(KB)$,
we build from the unit clauses a {\em prior} CPT for $A$ as its prior probability distribution.
(2) If $A$ only has non-unit clauses in ${\cal I}_{clause}(KB)$,
we build from the clauses a {\em posterior} CPT for $A$ as its posterior probability distribution.
(3) Otherwise, we prepare for $A$ both a prior CPT 
(from the unit clauses) and a posterior CPT (from the non-unit clauses).
In this case, $A$ is attached with the posterior CPT; the prior CPT for $A$ would be used,
if $A$ is a state variable, as the probability distribution of $A$ in time slice 0
(only in the case that a two-slice DBN is unrolled 
into a stationary DBN starting with time slice 0).
 
Assume that the parent nodes of $A$ are derived from $n$ ($n\geq 1$)
different influence clauses in ${\cal I}_{clause}(KB)$.
Suppose these clauses share the following CPTs 
in $T_x$: ${\bf P}(A_1|B_1^1,...,B_{m_1}^1)$, ..., and 
${\bf P}(A_n|B_1^n,...,B_{m_n}^n)$. (Recall that
an influence clause prefixed with a number $k$ shares the CPT
attached to the $k$-th clause in $PB$.)
Then the CPT for $A$ is computed by combining
the $n$ CPTs in terms of the combination rule $CR$
specified in Definition \ref{kb}. 

\begin{example}[Example \ref{aids-eg2} continued]
{\em
Let CPT$_i$ denote the CPT attached to the $i$-th clause in $PB_1$.
Consider the random variables in $2{\cal S}_{net}(KB_1)_{Q,E}$.
Since $aids(p1)$ has three parent nodes, derived from the 3rd and 4-th
clause in $PB_1$ respectively, the posterior CPT for $aids(p1)$ 
is computed by combining CPT$_3$ and CPT$_4$. $aids(p1)$  
has also a prior CPT, CPT$_1$, derived from the 1st clause in $PB_1$.
For the same reason, the posterior CPT for $aids(p2)$ 
is computed by combining CPT$_3$ and CPT$_4$.
The posterior CPT for $aids(p3)$ is CPT$_3$ and its 
prior CPT is CPT$_2$. $contact(p1,p2)$ and $contact(p2,p1)$
have only prior CPTs, namely CPT$_5$ and CPT$_6$.
Note that state input nodes, $aids(p1)_{t-1}$, $aids(p2)_{t-1}$ and $aids(p3)_{t-1}$,
do not need to have a CPT; they will be expanded, during the process of unrolling
the two-slice DBN into a stationary DBN, to cover the time slices involved in
the given query and evidence nodes.
If the resulting stationary DBN starts with time slice 0, 
the prior CPTs, CPT$_{aids(p1)_0}$ and CPT$_{aids(p3)_0}$, 
for $aids(p1)$ and $aids(p3)$ are used
as the probability distributions of $aids(p1)_0$ and $aids(p3)_0$.

Note that $aids(p2)$ is a state variable, but there is no unit influence clause 
available to build a prior CPT for it. 
We have two ways to derive a prior CPT, CPT$_{aids(p2)_0}$, for $aids(p2)$ 
from some existing CPTs. (1) CPT$_{aids(p2)_0}$ comes from
averaging CPT$_{aids(p1)_0}$ and CPT$_{aids(p3)_0}$.
For instance, let the probability of $aids(p1)=yes$ be $0.7$ in CPT$_{aids(p1)_0}$ 
and the probability of $aids(p3)=yes$ be $0.74$ in CPT$_{aids(p3)_0}$. Then 
the probability of $aids(p2)=yes$ is $(0.7+0.74)/2 = 0.72$ in CPT$_{aids(p2)_0}$.
(2) CPT$_{aids(p2)_0}$ comes from averaging the 
posterior probability distributions of $aids(p2)$. 
For instance, let $\{0.9,0.7,0.4,0.8\}$ be the posterior
probabilities of $aids(p2)=yes$ in the posterior CPT for $aids(p2)$. 
Then the probability of $aids(p2)=yes$ is 
$(0.9+0.7+0.4+0.8)/4 = 0.7$ in CPT$_{aids(p2)_0}$. 
}
\end{example}

\section{Related Work}
\label{sec-related-work}
A recent overview of existing representational frameworks that combine probabilistic
reasoning with logic (i.e. logic-based approaches) 
or with relational representations (i.e. non-logic-based approaches)
is given by De Raedt and Kersting \cite{deke2003}. 
Typical non-logic-based approaches include 
probabilistic relational models (PRM), which are based on the entity-relationship 
(or object-oriented) model \cite{getoor01,jae97,pk2000}, and 
relational Markov networks, which combine Markov networks
and SQL-like queries \cite{tak2002}. 
Representative logic-based approaches include
frameworks based on the KBMC (Knowledge-Based Model Construction) idea 
\cite{bac90,breese92,fl98,glesner95,gc93,Kersting2000,ngo-had97,poole93},
stochastic logic programs (SLP) 
based on stochastic context-free grammars \cite{cuss2000,mugg96},
parameterized logic programs based on distribution semantics (PRISM) \cite{SK2001}, and more.
Most recently, a unifying framework, called {\em Markov logic},
has been proposed by Domingos and Richardson \cite{dr2004}.
Markov logic subsumes first-order logic and Markov networks.
Since our work follows the KBMC idea focusing on 
how to build a Bayesian network directly from
a logic program, it is closely related to three representative existing 
PLP approaches: the context-sensitive PLP developed 
by Haddawy and Ngo \cite{ngo-had97},
Bayesian logic programming proposed 
by Kersting and Raedt \cite{Kersting2000}, and 
the time parameter-based approach presented 
by Glesner and Koller \cite{glesner95}.
In this section, we make a detailed comparison
of our work with the three closely related approaches.

\subsection{Comparison with the Context-Sensitive PLP Approach}
\label{sec-comp-ngo}
The core of the context-sensitive PLP is a probabilistic knowledge base (PKB). 
In order to see the main differences from our Bayesian knowledge base (BKB), 
we reformulate its definition here.

\begin{definition}
\label{pkb}
{\em
A {\em probabilistic knowledge base} is a four tuple
$<$$PD,PB, CB,CR$$>$, where
\begin{itemize}
\item
$PD$ defines a set of probabilistic predicates ({\em p-predicates})
of the form $p(T_1, ...,$ $T_m, V)$ where all arguments $T_i$s are
typed with a finite domain and the last argument $V$ takes on
values from a probabilistic domain $DOM_p$.

\item
$PB$ consists of {\em probabilistic rules} of the form
\begin{equation}
\label{p-rule}
P(A_0|A_1,...,A_l)=\alpha\leftarrow B_1,...,B_m, \neg  C_1, ..., \neg  C_n
\end{equation}
where $0\leq \alpha \leq 1$, the $A_i$s are p-predicates, and the $B_j$s and $C_k$s
are context predicates ({\em c-predicates}) defined in $CB$.
 
\item
$CB$ is a logic program, and both $PB$ and $CB$ are acyclic.

\item
$CR$ is a combination rule. 
\end{itemize} 
}
\end{definition}

In a probabilistic rule (\ref{p-rule}), 
each p-predicate $A_i$ is of the form $q(t_1,...,t_m,v)$, which simulates an equation 
$q(t_1,...,t_m)=v$ with $v$ being a value from the probabilistic domain of 
$q(t_1,...,t_m)$. For instance, let 
$D_{color} = \{red, green, blue\}$ be the probabilistic domain
of $color(X)$, then 
the p-predicate $color(X, red)$ simulates $color(X)=red$, 
meaning that the color of $X$ is $red$.  
The left-hand side $P(A_0|A_1,...,A_l)=\alpha$ expresses
that the probability of $A_0$ conditioned on $A_1,...,A_l$ is $\alpha$.
The right-hand side $B_1,...,B_m, \neg  C_1, ..., \neg  C_n$ is the {\em context}
of the rule where the $B_j$s and $C_k$s are c-predicates. Note that 
the sets of p-predicate and c-predicate symbols are disjoint. A separate
logic program $CB$ is used to evaluate the context of a probabilistic rule.
As a whole, the above probabilistic rule states that 
for each of its (Herbrand) ground instances
\[P(A_0'|A_1',...,A_l')=\alpha\leftarrow B_1',...,B_m', \neg  C_1', ..., \neg  C_n'\]
if the context $B_1',...,B_m', \neg  C_1', ..., \neg  C_n'$ is true in $CB$
under the program completion semantics, the probability of
$A_0'$ conditioned on $A_1',...,A_l'$ is $\alpha$.

PKB and BKB have the following important differences.

First, probabilistic rules of form (\ref{p-rule}) in PKB
contain both logic representation (right-hand side) 
and probabilistic representation (left-hand side)
and thus are not logic clauses. The logic part and the probabilistic part of a rule
are separately computed against $CB$ and $PB$, respectively. 
In contrast, BKB uses logic clauses of form (\ref{cll-2}), 
which naturally integrate the direct influence information, the context and 
the type constraints.
These logic clauses are evaluated against a single logic program 
$PB\cup CB$, while the probabilistic information
is collected separately in $T_x$. 

Second, logic reasoning in PKB relies
on the program completion semantics and is carried out
by applying SLDNF-resolution. But in BKB, logic inferences are based
on the well-founded semantics and are performed by applying SLG-resolution.
The well-founded semantics resolves the problem of inconsistency with the program completion semantics,
while SLG-resolution eliminates the problem of infinite loops with SLDNF-resolution.
Note that the key significance of BKB using the well-founded semantics lies in
the fact that a unique set of influence clauses can be derived, which lays
a basis on which both the declarative and procedural semantics for BKB are developed.

Third, most importantly PKB has no mechanism for handling cyclic influences. 
In PKB, cyclic influences are defined to be {\em inconsistent} 
(see Definition 9 of the paper \cite{ngo-had97}) and thus are excluded 
(PKB excludes cyclic influences by requiring its programs be acyclic).
In BKB, however, cyclic influences are interpreted as feedbacks, thus
implying a time sequence. This allows us to derive a stationary DBN
from a logic program with recursive loops.  

Recently, Fierens, Blockeel, Ramon and Bruynooghe \cite{fbrb2004} introduced
{\em logical Bayesian networks} (LBN). LBN is similar to PKB except that it 
separates logical and probabilistic information. That is, LBN converts rules
of form (\ref{p-rule}) into the form
\[A_0|A_1,...,A_l\leftarrow B_1,...,B_m, \neg  C_1, ..., \neg  C_n\]
where the $A_i$s are p-predicates with the last argument $V$ removed, and the $B_j$s and $C_k$s
are c-predicates defined in $CB$. This is not a standard clause 
of form (\ref{eq1}) as defined in logic programming \cite{Ld87}.
Like PKB, LBN differs from BKB in the following: 
(1) it has no mechanism for handling cyclic influences
(see Section 3.2 of the paper \cite{fbrb2004}), and (2)
although the well-founded semantics is also used for the
logic contexts, neither declarative nor procedural semantics
for LBN has been formally developed.

\subsection{Comparison with Bayesian Logic Programming}

Building on Ngo and Haddawy's work, Kersting and De Raedt \cite{Kersting2000}
introduce the framework of Bayesian logic programs.
A {\em Bayesian logic program} (BLP) is a triple $<$$P,T_x, CR$$>$
where $P$ is a well-defined logic program, $T_x$ consists of CPTs associated 
with each clause in $P$, and $CR$ is a combination rule. 
A distinct feature of BLP over PKB is its separation of 
probabilistic information ($T_x$) from logic clauses ($P$).
According to \cite{Kersting2000},
we understand that a  {\em well-defined} logic program is an acyclic positive 
logic program satisfying the range restriction.\footnote{A logic program is said
to be {\em range-restricted} if all variables appearing in the head of
a clause appear in the body of the clause.} 
For instance, a logic program containing clauses like 
$r(X)\leftarrow r(X)$ (cyclic) or $r(X)\leftarrow s(Y)$ (not range-restricted)
is not well-defined. BLP relies on the least Herbrand model 
semantics and applies SLD-resolution to make
backward-chaining inferences. 

BLP has two important differences from BKB.
First, it applies only to positive logic programs. Due to this, it cannot
handle contexts with negated atoms. (In fact, no contexts are 
considered in BLP.) Second, it does not allow cyclic influences.
BKB can be viewed as an extension of BLP with mechanisms
for handling contexts and cyclic influences in terms of the well-founded semantics.
Such an extension is clearly nontrivial.

\subsection{Comparison with the Time Parameter-Based Approach}

The time parameter-based framework (TPF) proposed 
by Glesner and Koller \cite{glesner95} is also a triple
$<$$P,T_x, CR$$>$, where $CR$ is a combination rule,
$T_x$ is a set of CPTs that are represented as decision trees,
and $P$ is a logic program with the property that each predicate contains
a time parameter and that in each clause 
the time argument in the head is at least one time
step later than the time arguments in the body.
This framework is implemented in Prolog, i.e. clauses are represented
as Prolog rules and goals are evaluated applying SLDNF-resolution.
Glesner and Koller \cite{glesner95} state: ``... In principle, this free variable $Y$
can be instantiated with every domain element. (This is the approach taken
in our implementation.)" By this we understand that they consider typed
logic programs with finite domains.

We observe the following major differences between TPF and BKB.
First, TPF is a temporal model and its logic programs contain a
time argument for every predicate. It always builds a DBN 
from a logic program even if there is no cyclic influence. 
In contrast, logic programs in BKB contain
no time parameters. When there is no cyclic influence, BKB
builds a regular Bayesian network from a logic program
(in this case, BKB serves as a non-temporal model); 
when cyclic influences occur, it builds a stationary DBN,
represented by a two-slice DBN (in this case, BKB serves as a 
special temporal model). Second, TPF uses time steps to describe 
direct influences (in the way that for any $A$ 
and $B$ such that $B$ has a direct influence on $A$,
the time argument in $B$ is at least one time step earlier than that in $A$),
while BKB uses time slices (implied by recursive loops of form (\ref{loop1}))
to model cycles of direct influences (feedbacks). 
Time-steps based frameworks like TPF are suitable to model flexible DBNs, 
whereas time-slices based approaches like BKB apply to
stationary DBNs. Third, most importantly TPF avoids recursive loops
by introducing time parameters to enforce acyclicity of a logic program. 
A serious problem with this method 
is that it may lose and/or produce wrong answers to some queries. To explain this,
let $P$ be a logic program and
$P_t$ be $P$ with additional time arguments added
to each predicate (as in TPF). 
If the transformation from $P$ to $P_t$ is
correct, it must hold that for any query $p(.)$ over $P$, 
an appropriate time argument $N=0,1,2,...$ can 
be determined such that the query
$p(., N)$ over $P_t$ has the same set of answers as $p(.)$ over $P$
when the time arguments in the answers are ignored.
It turns out, however, that this condition does not hold in general cases.
Note that finding an appropriate $N$ for a query $p(.)$ such that 
evaluating $p(., N)$ over $P_t$ (applying SLDNF-resolution) 
yields the same set of answers as evaluating $p(.)$ over $P$ 
corresponds to finding an appropriate depth-bound $M$ such that
cutting all SLDNF-derivations for the query $p(.)$  
at depth $M$ does not lose any answers to $p(.)$. 
The latter is the well-known loop problem
in logic programming \cite{BAK91}. Since the loop problem is 
undecidable in general, there is no algorithm for
automatically determining such a depth-bound $M$ (rep. a 
time argument $N$) for an arbitrary query $p(.)$ 
\cite{BAK91,shen001,shen-tocl}. We further illustrate
this claim using the following example.
\begin{example}
\label{eg-last}
{\em
The following logic program defines a $path$
relation; i.e. there is a path from $X$ to $Y$
if either there is an edge from $X$ to $Y$ or for some $Z$,
there is a path from $X$ to $Z$ and an edge from $Z$ to $Y$. 
\begin{tabbing}
$\qquad$ $P:\ $ \= 1. $\ e(s,b1).$\\
\>                 2. $\ e(b1,b2).$\\
\>                  $\qquad ......$\\
\>                 99. $\ e(b98,b99).$\\
\>                 100. $\ e(b99,g).$\\
\>                 101. $\ path(X,Y) \leftarrow e(X,Y).$\\
\>                 102. $\ path(X,Y) \leftarrow path(X,Z), e(Z,Y).$
\end{tabbing}

To avoid recursive loops, TPF may transform $P$ into the following
program.
\begin{tabbing}
$\qquad$ $P_t:\ $ \= 1. $\ e(s,b1,0).$\\
\>                   2. $\ e(b1,b2,0).$\\
\>                     $\qquad ......$\\
\>                 99. $\ e(b98,b99,0).$\\
\>                 100. $\ e(b99,g,0).$\\
\>                101. $\ e(X,Y,T1) \leftarrow T2=T1-1, e(X,Y,T2).$\\
\>                102. $\ path(X,Y,T1) \leftarrow T2=T1-1, e(X,Y, T2).$\\
\>                103. $\ path(X,Y,T1) \leftarrow T2=T1-1, path(X,Z,T2), e(Z,Y,T2).$
\end{tabbing}

$P_t$ looks more complicated than $P$. In addition to having time arguments
and time formulas, it has a new clause, the 101st clause, formulating that 
$e(X,Y)$ being true at present implies it is true in the future. 

Let us see how to check if there is a path from $s$ to $g$. In the original program
$P$, we simply pose a query $?-path(s,g)$. In the transformed program $P_t$,
however, we have to determine a specific time parameter $N$ and then pose a query
$?-path(s,g,N)$, such that evaluating $path(s,g)$ over $P$ yields the same answer
as evaluating $path(s,g,N)$ over $P_t$. Interested readers can practice this 
query evaluation using different values for $N$. The answer to $path(s,g)$ over $P$
is $yes.$ However, we would get an answer $no$ to the query $path(s,g,N)$ over $P_t$
if we choose any $N<100$. 
}   
\end{example}

\section{Conclusions and Discussion}
We have developed a novel theoretical framework for deriving a stationary DBN 
from a logic program with recursive loops.
We observed that recursive loops in a logic program imply a time sequence
and thus can be used to
model a stationary DBN without using explicit time parameters.  
We introduced a Bayesian knowledge 
base with logic clauses of form (\ref{cll-2}).
These logic clauses naturally integrate the direct influence information, the context and 
the type constraints, and are evaluated under the well-founded semantics. 
We established a declarative semantics for a Bayesian knowledge base
and developed algorithms that build 
a two-slice DBN from a Bayesian knowledge base.

We emphasize the following three points. 
\begin{enumerate}
\item
Recursive loops (cyclic influences) and
recursion through negation are unavoidable in modeling
real-world domains, thus the well-founded semantics together
with its top-down inference procedures is 
well suitable for the PLP application. 

\item 
Recursive loops define feedbacks, thus implying a time sequence. This
allows us to derive a two-slice DBN from a logic program containing
no time parameters. We point out, however, that the user is never required to 
provide any time parameters during the process of constructing such
a two-slice DBN. A Bayesian knowledge base defines a unique
space of random variables and a unique set of
influence clauses, whether it contains recursive loops or not.
From the viewpoint of logic, these random variables are ground atoms
in the Herbrand base; their truth values are determined by the well-founded model
and will never change over time.\footnote{ 
However, from the viewpoint of Bayesian networks the 
probabilistic values of these random variables (i.e. values from their probabilistic
domains) may change over time.} Therefore, a Bayesian network is built
over these random variables, independently of any time factors (if any).
Once a two-slice DBN has been built, the time intervals over it
would become clearly specified, thus the user can present queries and evidences over
the DBN using time parameters at his/her convenience.

\item
Enforcing acyclicity of a logic program by introducing
time parameters is not an effective way to handle recursive
loops. Firstly, such a method transforms the original non-temporal logic program
into a more complicated temporal program and builds a dynamic
Bayesian network from the transformed program even if there exist
no cyclic influences (in this case, there is no state variable
and the original program defines a regular Bayesian network). 
Secondly, it relies on time steps to
define (individual) direct influences, but recursive loops
need time slices (intervals) to model cycles of direct influences
(feedbacks). Finally, to pose a query over the transformed program,
an appropriate time parameter must be specified. As illustrated in Example \ref{eg-last},
there is no algorithm for automatically determining such a time parameter
for an arbitrary query.
\end{enumerate}

Promising future work includes (1) developing algorithms for learning
BKB clauses together with their CPTs from data and (2) 
applying BKB to model large real-world problems.
We intend to build a large Bayesian knowledge base for traditional Chinese medicine,
where we already have both a large volume of collected diagnostic rules
and a massive repository of diagnostic cases.

\section*{Acknowledgements}
We are grateful to 
several anonymous referees for their constructive comments,
which greatly helped us improve the presentation. 

\end{document}